\documentclass[10pt,journal,compsoc]{IEEEtran}

\ifCLASSOPTIONcompsoc
  \usepackage[nocompress]{cite}
\else
  \usepackage{cite}
\fi

\ifCLASSINFOpdf
\usepackage{graphicx}
\else
\fi

\usepackage{amsmath}
\usepackage{amsfonts}
\usepackage{algorithmic}

\usepackage{array}
\newcommand*{\aaagan}{A\textsuperscript{3}GAN}

\ifCLASSOPTIONcompsoc
  \usepackage[caption=false,font=footnotesize,labelfont=sf,textfont=sf]{subfig}
\else
  \usepackage[caption=false,font=footnotesize]{subfig}
\fi

\usepackage{url}

\usepackage{xcolor}
\usepackage{multirow}
\usepackage{multicol}
\usepackage{hhline}
\usepackage{booktabs}
\usepackage{tabularx}
\usepackage{anyfontsize}
\usepackage{threeparttable}
\usepackage{ragged2e}
\renewcommand{\raggedright}{\leftskip=0pt \rightskip=0pt plus 0cm}

\newcolumntype{s}{>{\hsize=0.7\hsize}>{\centering\arraybackslash}X}
\newcolumntype{b}{>{\hsize=1.05\hsize}>{\raggedright\arraybackslash}X}
\newcolumntype{L}{>{\raggedright\arraybackslash}p{2em}}
\newcommand{\ra}[1]{\renewcommand{\arraystretch}{#1}}

\begin{document}
\title{\aaagan: An Attribute-aware Attentive \\ Generative Adversarial Network for Face Aging}
%
%
%
\author{Yunfan~Liu,~Qi~Li,~Zhenan~Sun,~\IEEEmembership{Senior Member,~IEEE},~and~Tieniu~Tan,~\IEEEmembership{Fellow,~IEEE}%
\IEEEcompsocitemizethanks{\IEEEcompsocthanksitem Y. Liu is with the Center for Research on Intelligent Perception and Computing, National Laboratory of Pattern Recognition, Institute of Automation, Chinese Academy of Sciences, Beijing 100190, China and also with University of Chinese Academy of Sciences, Beijing 100190, China. E-mail: yunfan.liu@cripac.ia.ac.cn.%
\IEEEcompsocthanksitem Q. Li, Z. Sun, and T. Tan are with the Center for Research on Intelligent
Perception and Computing, National Laboratory of Pattern Recognition, Institute of Automation, CAS Center for Excellence in Brain Science and Intelligence Technology, Chinese Academy of Sciences, Beijing 100190, China. E-mail: \{qli, znsun, tnt\}@nlpr.ia.ac.cn.}
}


\IEEEtitleabstractindextext{%
\raggedright{
\begin{abstract}
Face aging, which aims at aesthetically rendering a given face to predict its future appearance, has received significant research attention in recent years.
Although great progress has been achieved with the success of Generative Adversarial Networks (GANs) in synthesizing realistic images, most existing GAN-based face aging methods have two main problems:
1) unnatural changes of high-level semantic information (e.g.~facial attributes) due to the insufficient utilization of prior knowledge of input faces, and
2) distortions of low-level image content including ghosting artifacts and modifications in age-irrelevant regions.
In this paper, we introduce~\aaagan, an \textbf{A}ttribute-\textbf{A}ware \textbf{A}ttentive face aging model to address the above issues.
Facial attribute vectors are regarded as the conditional information and embedded into both the generator and discriminator, encouraging synthesized faces to be faithful to attributes of corresponding inputs.
To improve the visual fidelity of generation results, we leverage the attention mechanism to restrict modifications to age-related areas and preserve image details.
Moreover, the wavelet packet transform is employed to capture textural features at multiple scales in the frequency space.
Extensive experimental results demonstrate the effectiveness of our model in synthesizing photorealistic aged face images and achieving state-of-the-art performance on popular face aging datasets.
\end{abstract}}

}

\maketitle

\IEEEdisplaynontitleabstractindextext

\IEEEraisesectionheading{\section{Introduction}\label{sec:introduction}}

\IEEEPARstart{F}{ace} aging, also known as age progression, refers to rendering a given face image with realistic aging effects while still preserving personalized features~\cite{ramanathan2009age,fu2010age,yang2017learning}.
Applications of face aging techniques range from social security to digital entertainment, including predicting the contemporary appearance of lost individuals or wanted suspects based on outdated photos and bringing improvements to face recognition systems in the cross-age verification scenario.
Because of the significant practical value, face aging has received considerable research attention but remains challenging due to its intrinsic complexity.

In the last two decades, face aging has witnessed impressive progress and a large number of approaches have been proposed to address this problem, which could be generally divided into two categories: physical model-based methods~\cite{todd1980perception,lanitis2002toward,golovinskiy2006statistical,ramanathan2008modeling,suo2010compositional,tazoe2012facial} and prototype-based methods~\cite{burt1995perception,tiddeman2001prototyping,kemelmacher2014illumination,shu2016kinship,tang2017personalized}.
Physical model-based methods simulate the profile growth mechanically via parameterized models of facial shape and texture, while prototype-based methods render aging effects by applying learned translation patterns between prototype faces (averaged faces of pre-defined age groups) to the test image.
Although obvious aging signs could be synthesized using these two types of methods, they suffer from extremely high computational expenses and limited generalization ability~\cite{yang2017learning}.

In recent years, with the great success of Generative Adversarial Networks (GANs)~\cite{goodfellow2014generative} in image synthesis and translation tasks, many studies resort to GAN-based frameworks to solve the face aging problem~\cite{zhang2017age,li2018global,wang2018face_aging,yang2017learning}.
These methods model the mapping function between distributions of young and old face images and directly translate test faces into the target age group via learned mappings.
The most remarkable advantage of GAN-based methods over previous conventional approaches (physical model-based and prototype-based methods) is that synthesized face images are much more visually plausible and have fewer ghosting artifacts.
Moreover, GAN-based models could be trained in an end-to-end manner, which significantly reduces the overall complexity of the algorithm.

\begin{figure*}[t]
\begin{center}
\includegraphics[width=1.0\linewidth]{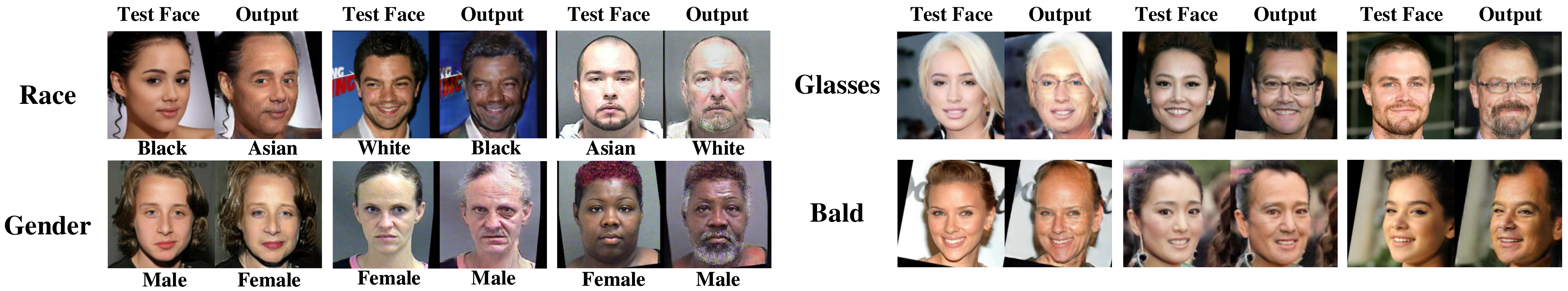}
\end{center}
\caption{Examples of face aging with mismatched facial attributes generated by face aging model without facial attribute embedding. Four attributes (Race, Gender, Glasses, and Bald) are considered and three sample results are presented for each. Labels of `Race' and `Gender' are all obtained via the public face analysis API of Face++~\cite{face2018toolkit} and placed under each image.}
\label{fig:faceAttIdt}
\end{figure*}

Since multiple face images of the same subject at different ages are prohibitively expensive to collect in practice, most GAN-based methods resort to unpaired face aging data to train the model.
However, these approaches mainly focus on simulating mappings between image contents while neglecting other critical semantic conditional information of the input (e.g., facial attributes), and thus fail to regulate the training process accordingly.
Concretely, a given young face image might map to multiple elderly face candidates in unpaired scenarios, which may mislead the model to establish translation patterns other than aging if no high-level conditional information is considered.
Consequently, serious ghosting artifacts and even incorrect facial attributes may appear in synthesized face images, which seriously reduces the authenticity and rationality of generation results.
For example, Fig.~\ref{fig:faceAttIdt} shows several face aging results with mismatched attributes.
In the rightmost face aging result under `Gender', the beard is mistakenly attached to the input female face image, which is almost impossible to happen in the natural aging process.
This is because the model learns that growing a beard is a typical sign of aging, but fails to recognize that this does not happen to a woman since no conditional information of the test face is involved in the training process.

In order to preserve personalized characteristics of input faces, many recent face aging studies attempt to regulate generation results by enforcing the identity consistency~\cite{zhang2017age,li2018global,wang2018face_aging,yang2017learning}.
However, as shown in Fig.~\ref{fig:faceAttIdt}, the identity of the test face is well preserved in the output for all sample results, nevertheless, unnatural changes of facial attributes could still be observed.
This suggests that well-maintained identity information does NOT imply reasonable aging results when training with unpaired data.
Therefore, merely enforcing identity consistency is insufficient to eliminate matching ambiguities, and thus fails to achieve satisfactory face aging performance in unpaired training scenarios.

In addition to undesired changes of facial attributes, another critical problem of existing GAN-based face aging method is that image contents irrelevant to age progression (e.g., image background) are not well preserved in the output, resulting in obvious ghosting artifacts and color distortions.
Basically, from the perspective of conditional image translation, face aging could be considered as adding representative signs of aging (e.g., wrinkles, eye bags, and laugh lines) to the input face image.
Therefore, image modifications are supposed to be restricted to those regions highly relevant to age changes and image contents should be well preserved elsewhere.
However, most existing GAN-based face aging methods do not enforce the constraint on regions of modification, instead, the pixel at each spatial location of the synthesis result is re-estimated by the generator.
Consequently, unintended correspondences between image contents other than age translation (e.g.~clothes and accessories) would be inevitably established, which heavily increases the chance of introducing age-irrelevant image modifications and ghosting artifacts.

To solve the above-mentioned issues, in this paper, we propose \aaagan, a GAN-based framework for \textbf{A}ttribute-\textbf{A}ware \textbf{A}ttentive face aging.
Different from existing methods in the literature, we involve semantic conditional information of the input by embedding facial attribute vectors into both the generator and discriminator, so that the model could be guided to output elderly face images with attributes faithful to the corresponding input.
To improve the visual quality of synthesized face images, we leverage the attention mechanism to restrict modifications to age-related areas and preserve details in input images.
Furthermore, to enhance aging details, based on the observation that signs of aging are mainly represented by wrinkles, eye bags, and laugh lines, which could be treated as local textures, we employ wavelet packet transform in the critic network to extract features at multiple scales in the frequency space efficiently.

Main contributions of this study are summarized as follows:
\begin{itemize}
\item An effective end-to-end GAN-based network, \aaagan, is proposed to solve the face aging problem. Specifically, facial attributes are embedded as the semantic conditional information into both the generator and discriminator to enforce more fine-grained consistency between input and generation results.
Besides, a wavelet packet transform module is adopted to extract features of aging textures at multiple scales in the frequency domain for generating more realistic details of aging effects.

\item To improve the quality of synthesized images and suppress ghosting artifacts, attention mechanism is introduced to help restrict image modifications to age-related image regions.

\item Extensive experiments have been conducted to demonstrate the ability of the proposed method in rendering accurate aging effects and preserving information of both identity and facial attributes. Quantitative comparison with other four advanced face aging benchmarks indicates that our method achieves state-of-the-art performance.
\end{itemize}

Compared to our previous work in~\cite{Liu2019Attribute}, this paper has the following extensions:
1) attention mechanism is introduced to help improve the visual quality of generation results via restricting modifications to image regions closely related to age progression;
2) generalization ability of the proposed method is investigated by testing the trained model on other widely used face datasets, including FG-NET~\cite{FGNET} and CelebA~\cite{liu2015faceattributes}.
3) we refined our model to obtain better results, and a thorough comparison with four GAN-based benchmark methods is provided to demonstrate the effectiveness of the proposed model in achieving reasonable and lifelike face aging results.

\begin{table*}[ht]
\ra{1.1}
\centering
\caption{Comparison between our model and previous GAN-based face aging methods.}
\begin{tabularx}{\textwidth} {@{} sbbb @{}}
\toprule
\multicolumn{1}{c}{Method} & \multicolumn{1}{c}{Main Features} & \multicolumn{1}{c}{Evaluation Metrics} & \multicolumn{1}{c}{Remarks} \\
\midrule
Conditional Adversarial Autoencoder (CAAE)~\cite{zhang2017age} & Age and identity translation are achieved by traversing on a low-dimensional manifold $\mathcal{M}$. & Identity permanence and visual fidelity (user study). & Only subtle aging textures are generated, which are insufficient to reflect associated age changes.\\
\midrule
Global and Local Consistent Age GAN (GLCA-GAN)~\cite{li2018global} & Three face patches are translated via dedicated sub-networks besides the global generator. & Face verification on age progression/regression (LightCNN-29~\cite{wu2018light}) & Extra network structures in the generator and accurate face patch cropping are required.\\
\midrule
Identity Preserved Conditional GAN (IPCGAN)~\cite{wang2018face_aging} & A pre-trained AlexNet~\cite{krizhevsky2012imagenet} is adopted to preserve the identity information in the feature space. & Face verification and age classification (user study); Inception Score (feature extracted by VGG~\cite{Simonyan14c}). & Training and testing face images are in low resolution ($128\times 128$).\\
\midrule
Pyramid-Structured Discriminator GAN (PSD-GAN)~\cite{yang2017learning} & A VGG-16 network~\cite{Simonyan14c} is used to extract multi-level age-related features for discrimination. & Age estimation and face verification (public face analysis tools of Face++~\cite{face2018toolkit})& The VGG-16 network in the discriminator requires pre-training on an age estimation task.\\
\midrule
\midrule
Attribute-aware Attentive GAN (\aaagan, our model) & Facial attributes are involved as conditional semantic information, and attention mechanism is adopted to further refine image quality. & Age estimation, face verification, and attribute preservation (public face analysis tools of Face++~\cite{face2018toolkit}) & Wavelet packet transform is employed in the discriminator to compute multi-scale textural features in the frequency space.\\
\bottomrule
\end{tabularx}
\label{table:MethodCompare}
\end{table*}

The rest of this paper is organized as follows:
Section~\ref{sec:related_works} briefly reviews related works on face aging and attention mechanism.
Detailed descriptions of the proposed method is provided in Section~\ref{sec:the_proposed_method}.
Experimental results are reported in Section~\ref{sec:experiments} presents.
Section~\ref{sec:conclusion} concludes this work and discusses possible future research directions.

\section{Related Works}\label{sec:related_works}
\subsection{Face Aging}
In the last few decades, face aging has been a very popular research topic and a great number of algorithms have been proposed to solve this problem.
In general, these methods could be divided into two categories: physical model-based methods and prototype-based methods.

Physical model-based methods are the initial explorations of face aging and they simulate changes of facial appearance w.r.t. time by modeling both geometrical and textural features of human faces.
As one of the earliest attempts, Todd et al.~\cite{todd1980perception} model the profile growth via the revised cardioidal strain transformation.
Subsequent works investigate the problem from various biological aspects including muscles and overall facial structures~\cite{lanitis2002toward,golovinskiy2006statistical,ramanathan2008modeling,suo2010compositional,tazoe2012facial}.
However, physical model-based algorithms are computationally expensive and difficult to generalize as they heavily depend on specific empirical aging rules.

\begin{figure*}[ht]
\begin{center}
\includegraphics[width=0.95\linewidth]{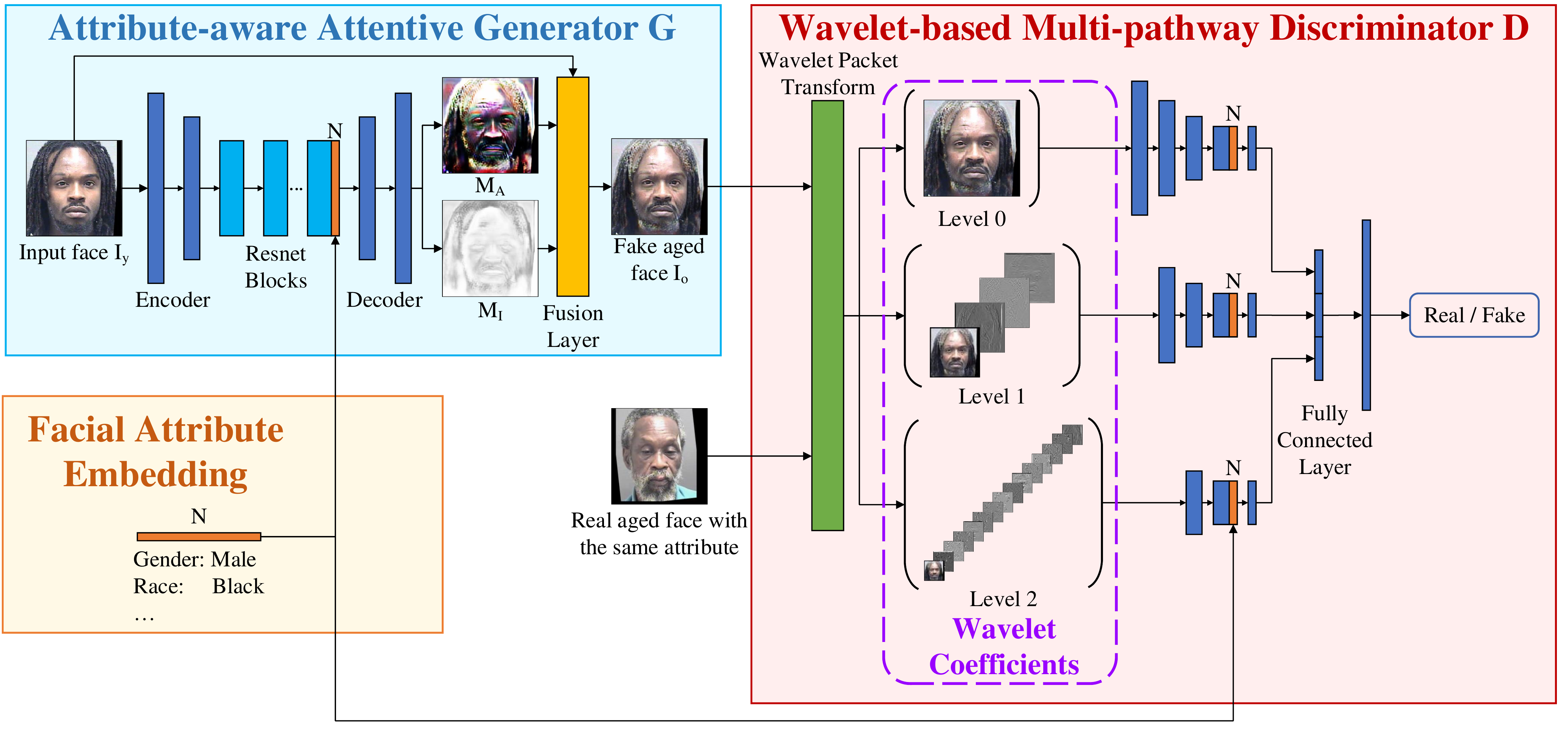}
\end{center}
\caption{An overview of the proposed~\aaagan model. An hourglass-shaped generator G learns the age mapping and outputs lifelike elderly face images. A discriminator D is employed to distinguish synthesized face images from generic ones, based on multi-scale wavelet coefficients computed by the wavelet packet transform module. The N-dimensional attribute vector describing the input face image is embedded to both the generator and discriminator to reduce matching ambiguity inherent to unpaired training data.}
\label{fig:framework}
\end{figure*}

As for data-driven prototyping approaches, Burt et al.~\cite{burt1995perception} propose to divide faces into age groups, each represented by an average face (the prototype), and regard differences between average faces as aging patterns. Following~\cite{burt1995perception}, many prototype-based methods are proposed to improve the face aging result~\cite{tiddeman2001prototyping,kemelmacher2014illumination,shu2016kinship,tang2017personalized}.
However, the main problem of prototype-based methods is that personalized features are eliminated when calculating averaged faces thus the identity information is not well preserved in aging results.
Moreover, since the pattern of age progression is largely determined by the averaged face of each age group, the diversity of visual effects of face aging is quite limited.

With the rapid development of deep learning theory, deep generative models with temporal architectures are proposed to model age progression with hierarchically learned representations~\cite{wang2016recurrent,duong2016longitudinal,duong2017temporal}.
However, in most of these works, face image sequence over a long age span for each subject is required thus their potential in practical application is limited.

Recently, Generative Adversarial Networks (GANs)~\cite{goodfellow2014generative} have achieved remarkable success in generating visually plausible images, and many efforts have been made to solve the problem of face aging taking the advantage of adversarial learning~\cite{zhang2017age,yang2017learning,li2018global,wang2018face_aging}.
Zhang et al.~\cite{zhang2017age} propose a conditional adversarial autoencoder (CAAE) to achieve age progression and regression by traversing in a low-dimensional feature manifold.
Li et al.~\cite{li2018global} attend to three manually selected facial patches where age effects are likely to appear, and separate generators are adopted to model the appearance change in these areas.
By incorporating a conditional age vector, Wang et al.~\cite{wang2018face_aging} achieve age progression to multiple target age groups with a single model.
Using a pre-trained deep model in the discriminator network, Yang et al.~\cite{yang2017learning} propose a GAN-based framework with pyramid-structured discriminator (PSD-GAN) to render aging effects.
A comprehensive summary of GAN-based face aging methods and comparisons between our method and previous state-of-the-art is provided in TABLE.~\ref{table:MethodCompare}.

\subsection{Attention Mechanism}
Attention plays an important role in the human visual system as it serves as a high-level understanding of the scene and could guide the bottom-up processing of detailed objects~\cite{itti1998model,rensink2000dynamic,corbetta2002control}.
In recent years, numerous attempts have been made to embed the attention mechanism into deep neural networks to improve the performance.
Attention mechanism has been successfully applied to recurrent neural networks (RNN) and long short-term memory (LSTM) to tackle problems with sequential input, including neural machine translation~\cite{bahdanau2014neural,Vaswani2017attention}, visual question answering~\cite{xu2016ask,yang2016stacked,yu2017multi}, and caption generation~\cite{xu2015show}.

As for vision-related tasks, attention mechanism could be naturally introduced to guide the model to focus on specific image regions closely related to the target task. Wang et al.~\cite{wang2017residual} propose a residual attention network which could generate attention-aware features for image classification. Woo et al.~\cite{woo2018cbam} explore the effectiveness of a light-weight general attention module, Convolutional Block Attention Module (CBAM), in improving the performance of deep models in various vision tasks. Albert et al.~\cite{pumarola2018ganimation} adopt the spatial attention mechanism to synthesize face images with target expression. Attention is also widely used in solving image captioning~\cite{chen2017sca} and saliency detection problems~\cite{zhang2018progressive,liu2018picanet,chen2018reverse,wang2019salient,zhao2019pyramid}.

\section{The Proposed Method}\label{sec:the_proposed_method}

\subsection{Overview of the Framework}
In an unpaired face aging dataset, a given young face image might map to many elderly face candidates during the training process, which may mislead the model into learning translations other than aging if no conditional information is considered.
To solve this problem, we present a GAN-based face aging model that takes both young face images and their semantic information (i.e. facial attributes) as input and outputs visually plausible aged faces with consistent facial attributes.

Our model mainly consists of two key components: an attribute-aware attentive generator $G$ and a wavelet-based multi-pathway discriminator $D$.
The generator $G$ takes a young face image $I_y\in \mathbb{R}^{H\times W\times C}$ as input and predict the corresponding aged face $I_o$, while the discriminator $D$ encourages generation results to be indistinguishable from generic face images.
Unlike most existing face aging methods, the attribute of the input (denoted as $\alpha\in \mathbb{R}^{N}$, where $N$ is the number of facial attributes to be preserved) is considered as the conditional information and embedded into both $G$ and $D$ to ensure the attribute consistency.
An overview of the proposed framework is shown in Fig.~\ref{fig:framework}.

\subsection{Attribute-aware Attentive Generator}
Most existing GAN-based face aging methods~\cite{zhang2017age,li2018global,yang2017learning,wang2018face_aging} have two main problems:
\begin{enumerate}
\item Only images of young faces are taken as input to learn mappings between age groups, regardless of any prior knowledge that may have an influence on the visual pattern of age progression. Although constraints on identity information and pixel values are usually adopted to restrict modifications made to input images, facial attributes may still undergo unnatural translations (as shown in Fig.~\ref{fig:faceAttIdt}).

\item Although signs of aging concentrate on certain facial regions which only take up a small percentage of the entire image, pixel at each spatial location is re-estimated in the generation result. Consequently, unintended correspondences between image contents other than age translation (e.g.~background textures) would be inevitably established, which increases the chance of introducing age-irrelevant changes and ghosting artifacts.
\end{enumerate}

\begin{figure}[t]
\centering
\includegraphics[width=0.85\linewidth]{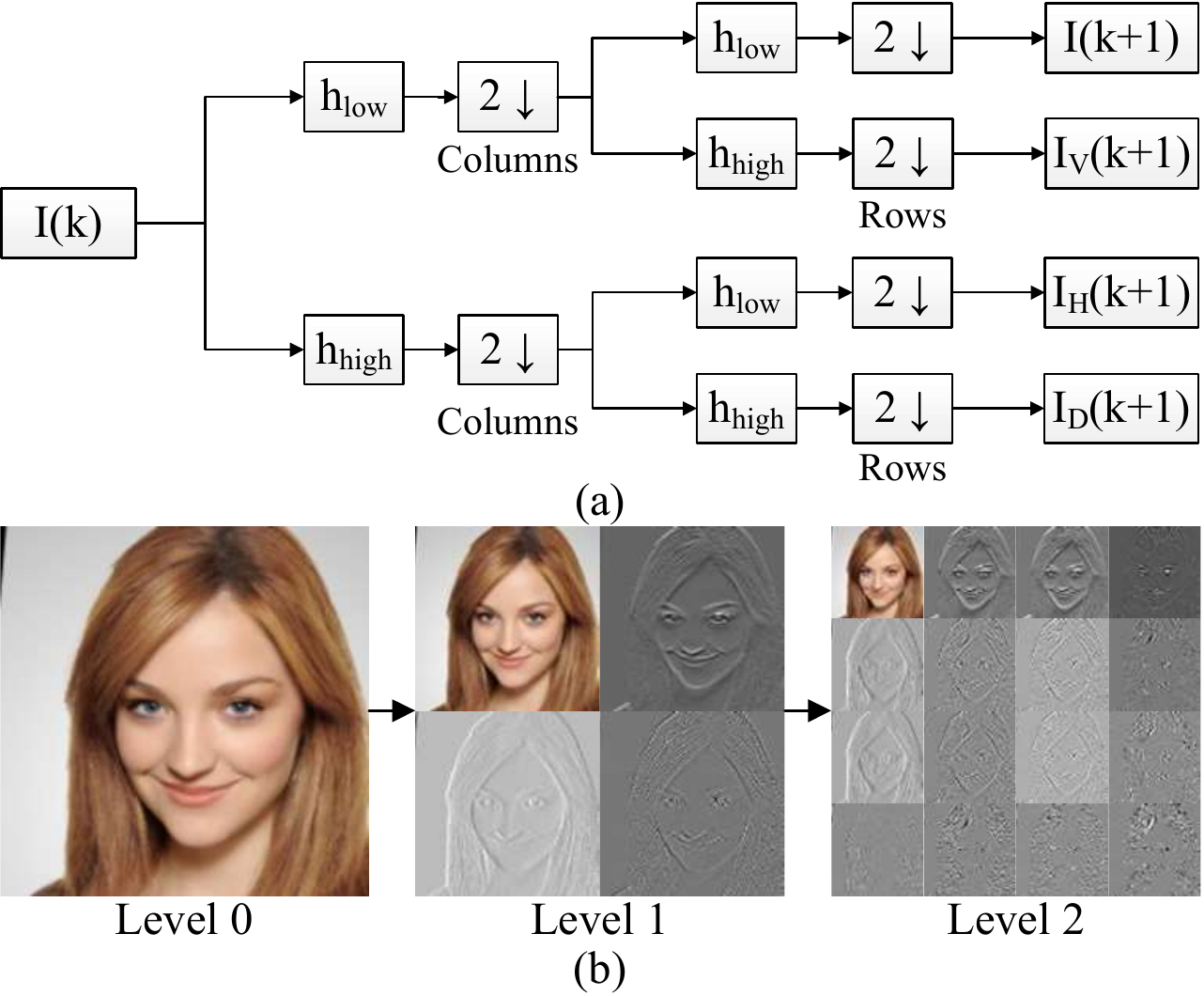}
\caption{Demonstration of wavelet packet transform. (a) Low-pass and high-pass decomposition filters ($h_{low}$ and $h_{high}$) are applied iteratively to the input on $k$-th level to compute wavelet coefficients on the next level; (b) a sample face image with its wavelet coefficients at different decomposing levels.}
\label{fig:waveletPacketTransform}
\end{figure}

To solve these problems, in this paper, we propose an attribute-aware attentive generator $G$ to achieve fine-grained face aging (detailed structure is shown in Table~\ref{table:structure_generator}).
We employ an hourglass-shaped fully convolutional network as the backbone of the generator, which has achieved success in previous image translation studies~\cite{johnson2016perceptual,zhu2017unpaired}.
Specifically, it consists of three key components: an encoder network, a decoder network, and six residual blocks in between as the bottleneck.
Unlike previous works, we propose to incorporate both low-level image data (pixel values) and high-level semantic information (facial attributes) into the face aging model to regulate image translation patterns and reduce the ambiguity of mappings between unpaired young and aged faces.
Concretely, the input facial attribute vector is replicated along the spatial dimension and then concatenated with the output of the last residual block (ResBlock6), as they both contain high-level representations of the input image.

Considering the fact that age progression is essentially the gradual emergence of aging signs, we naturally introduce the attention mechanism to guide the generator to concentrate more on image regions that aging signs are likely to appear.
This is achieved by estimating an attention mask describing the contribution of each pixel in the input image $I_y$ to the final generation result.
As shown in Fig.~\ref{fig:framework}, the decoder network outputs two feature maps, an attention mask $M_A\in [0,1]^{H\times W\times C}$ and an image map $M_I\in \mathbb{R}^{H\times W\times C}$.
These two feature maps are then fed into a fusion layer along with the input image $I_y$ to obtain the aged face image $I_o$, which could be formulated as,
\begin{equation}
I_o = M_A\odot I_y + (1-M_A)\odot M_I
\end{equation}
where $\odot$ denotes element-wise product and $M_A$ is replicated along the channel dimension to match the size of $M_I$ and $I_y$.

Intuitively, the attention mask $M_A$ indicates the proportion of the original input image being retained at each spatial location, or in other words, to what extent the image map $M_I$ contributes to the final output.
For example, as shown in Fig.~\ref{fig:framework}, brighter areas in $M_A$ suggest those regions of the final output $I_o$ retain more information from the input $I_y$, which are precisely image contents irrelevant to age changing (e.g.~background and clothes).
On the other hand, darker areas in $M_A$ refer to regions in $I_o$ that are more closely related to the image map $M_I$, that is, representative signs of face aging (i.e. forehead, hair, and laugh lines as shown in Fig.~\ref{fig:framework}).
The greatest advantage of adopting attention mechanism is that the generator could be guided to focus only on rendering age changing effects within specific regions, and pixels irrelevant to age progression could be directly obtained from the original input, resulting in more fine-grained image details with fewer ghosting artifacts.

\begin{table}[!t]
\caption{Architecture of the generator $G$}
\label{table:structure_generator}
\centering
\begin{threeparttable}
\begin{tabular}{@{}cccc@{}}
\toprule
Module                      & Layer      & K \:/\: S \:/\: P\tnote{*}  & Output Size \\
\midrule
\multirow{3}{*}{Encoder}    & Conv1      & $7\times7$ \:/\: 1 \:/\: 3  & $256\times256\times64$  \\
                            & Conv2      & $4\times4$ \:/\: 2 \:/\: 1  & $128\times128\times128$ \\
                            & Conv3      & $4\times4$ \:/\: 2 \:/\: 1  & $64\times64\times256$   \\
\midrule
\multirow{6}{*}{Bottleneck} & ResBlock1  & $3\times3$ \:/\: 1 \:/\: 1  & $64\times64\times256$ \\
                            & ResBlock2  & $3\times3$ \:/\: 1 \:/\: 1  & $64\times64\times256$ \\
                            & ResBlock3  & $3\times3$ \:/\: 1 \:/\: 1  & $64\times64\times256$ \\
                            & ResBlock4  & $3\times3$ \:/\: 1 \:/\: 1  & $64\times64\times256$ \\
                            & ResBlock5  & $3\times3$ \:/\: 1 \:/\: 1  & $64\times64\times256$ \\
                            & ResBlock6  & $3\times3$ \:/\: 1 \:/\: 1  & $64\times64\times256$ \\
\midrule
\multirow{2}{*}{Decoder}    & up~$\uparrow2$\:\&\:Conv1\tnote{$\dagger$}  & $3\times3$ \:/\: 1 \:/\: 1  & $128\times128\times128$ \\
                            & up~$\uparrow2$\:\&\:Conv2  & $3\times3$ \:/\: 1 \:/\: 1  & $256\times256\times64$  \\
\midrule
\multicolumn{2}{c}{Conv for estimating $M_A$}            & $7\times7$ \:/\: 1 \:/\: 3  & $256\times256\times1$  \\
\midrule
\multicolumn{2}{c}{Conv for estimating $M_I$}            & $7\times7$ \:/\: 1 \:/\: 3  & $256\times256\times3$  \\
\bottomrule
\end{tabular}
\begin{tablenotes}\footnotesize
\item[*] K, S, P denotes the size of kernel, stride, and padding, respectively.
\item[$\dagger$] up~$\uparrow2$\:\&\:Conv denotes upsampleing layer (scale factor of 2) followed by a convolutional layer
\end{tablenotes}
\end{threeparttable}
\end{table}

\subsection{Wavelet-based Multi-pathway Discriminator}
In face aging tasks, a discriminator network $D$ is introduced to distinguish synthetic aged face images from generic ones, and the generator learns to confuse $D$ with outputs of high visual fidelity.

In order to generate more accurate and lifelike aging details, Yang et al.~\cite{yang2017learning} exploit a deep network with VGG-16 structure~\cite{Simonyan14c} pre-trained on an age classification task to extract age-related features conveyed by faces.
Although multi-scale representations could be obtained, storing and forwarding through a deep network would damage the efficiency of the model.
Besides, pre-training also requires extra effort and might potentially limit the generalizability of the model due to the bias towards training dataset.

To overcome this issue, since typical signs of aging, e.g.~wrinkles, laugh lines, and eye bags, could be regarded as local image textures, we adopt wavelet packet transform (WPT) to transform the input image to the frequency domain and capture textural features.
Specifically, multi-level WPT (see Fig.~\ref{fig:waveletPacketTransform}) is performed to provide a more comprehensive analysis of textures at multiple scales in the given image.
Compared to extracting multi-scale features using a sequence of convolutional layers as in~\cite{yang2017learning}, the advantage of using WPT is that the computational cost is significantly reduced since wavelet coefficients could be calculated by simply forwarding through a single convolutional layer.
Therefore, WPT greatly reduces the number of convolutions performed in each forwarding process.
Although this part of the model has been simplified in terms of network structure, it still takes the advantage of multi-scale image texture analysis, which helps improve the visual fidelity of generated images.

The overall structure of $D$ is shown in Tabel~\ref{table:structure_discriminator}.
Specifically, WPT is applied to input images to perform wavelet decomposition, and coefficients at each decomposing level are concatenated along the channel dimension before fed into separate convolutional pathways for feature extraction.
To make $D$ gain the ability to tell whether attributes are preserved in generated images, the attribute vector $\alpha_y$ is also replicated and concatenated to the output of an intermediate convolutional block of each pathway.
At the end of $D$, same-sized outputs of all pathways are fused to form a single tensor and then fed into a fully connected network to produce the final rating score for the authenticity of input images.

\begin{table}[!t]
\caption{Architecture of the discriminator $D$}
\label{table:structure_discriminator}
\centering
\begin{threeparttable}
\begin{tabular}{@{}cl|l|l@{}}
\toprule
Pathway index               & \multicolumn{1}{c}{1}  & \multicolumn{1}{c}{2}   & \multicolumn{1}{c}{3} \\
\midrule
Input size                  & $256\times256\times3$  & $128\times128\times12$  & $64\times64\times48$  \\
\midrule
\multirow{7}{*}{\shortstack[c]{Convolutional\\ Pathway\tnote{*}}} \
                            & Conv-64   & -         & -        \\
                            & Conv-128  & Conv-128  & -        \\
                            & Conv-256  & Conv-256  & Conv-256 \\
                            & (concat $I_y$)  & (concat $I_y$)  & (concat $I_y$) \\
                            & Conv-512  & Conv-512  & Conv-512 \\
                            & Conv-512  & Conv-512  & Conv-512 \\
                            & Conv-1    & Conv-1    & Conv-1   \\
\midrule
FC Layer                    & \multicolumn{3}{c}{$4\times4\times3\rightarrow1$} \\
\bottomrule
\end{tabular}
\begin{tablenotes}\footnotesize
\item[*] For all layers in the convolutional pathway, the size of kernel, stride, and padding are set to 4, 2, 1, respectively. Only the number of channels of output for each layer is listed in the table.
\end{tablenotes}
\end{threeparttable}
\end{table}

\subsection{Objective Functions and Training Procedure}
The training objective of the proposed model consists of three parts:
an \textit{adversarial loss} to encourage the distribution of generated images to be indistinguishable from that of real images,
an \textit{identity loss} to preserve personalized characteristics of the input image,
and a \textit{pixel-level loss} to reduce the gap between the input and output of the generator in the image space.

\subsubsection{Adversarial Loss}
The adversarial process between the generator $G$ and discriminator $D$ encourages synthetic results to be photo-realistic and indistinguishable from real ones.
Besides visual fidelity, attribute consistency is also guaranteed by involving the attribute of input face images as conditional information in the adversarial process.

To achieve these two goals, unlike existing face aging methods, our discriminator network D is designed to take pair-wise input, i.e. aged face images and their corresponding attributes.
Our goal is to make $D$ gain the ability to discriminate generated aged face images from real ones and telling whether the input face image contains the desired attribute.
Therefore, to train the discriminator network, data pairs of real aged faces with attributes same as $I_y$, denoted by $\{I_o, \alpha\}$, are considered as positive samples.
Negative samples include image pairs of synthesized aged faces $G(I_y, \alpha)$ and their attributes $\alpha$, i.e.~$\{G(I_y, \alpha), \alpha\}$, as well as images pairs of real aged faces and mismatched attributes, i.e.~$\{I_o, \bar{\alpha}\}$.

Formally, the objective function for training the discriminative network $D$ consists of two parts, that is, $\mathcal{L}_{adv\_att}$ for checking the attribute consistency and $\mathcal{L}_{adv\_auth}$ for image authenticity discrimination.
Therefore, the adversarial loss could be formulated as,
\begin{equation}
\mathcal{L}_{adv_D} = \lambda_{att}\: \mathcal{L}_{adv\_att} + \mathcal{L}_{adv\_auth}
\end{equation}
where the parameter $\lambda_{att}$ controls the relative importance between $\mathcal{L}_{adv\_att}$ and $\mathcal{L}_{adv\_auth}$, which is initialized as 0 and then linearly increased during the training process.
This enables $D$ to firstly focus on discriminating fake images from real ones and then gradually adapts to the task of checking attribute consistency, which is critical for stabilizing the training process.
We follow WGAN~\cite{Arjovsky2017WGAN} and use the Wasserstein distance to measure the discrepancy between two data distributions.
Therefore, $\mathcal{L}_{adv\_att}$ and $\mathcal{L}_{adv\_auth}$ are formulated as follows,
\begin{align}\label{loss_adv_att}
\mathcal{L}_{adv\_att} = - &\: \mathbb{E}_{(I_o,\alpha)\sim P_o(I,\alpha)}\left[D(I_o,\alpha)\right]\nonumber\\
                         + &\: \mathbb{E}_{(I_o,\alpha)\sim P_o(I,\alpha)}\left[D(I_o,\bar\alpha)\right]
\end{align}
\begin{align}\label{loss_adv_real}
\mathcal{L}_{adv\_auth} = - &\: \mathbb{E}_{(I_o,\alpha)\sim P_o(I,\alpha)}\left[D(I_o,\alpha)\right]\nonumber\\
                          + &\: \mathbb{E}_{(I_y,\alpha)\sim P_y(I,\alpha)}\left[D(G(I_y,\alpha),\alpha)\right]
\end{align}
where $P_y$ and $P_o$ stand for the distribution of generic face images of young and old subjects, respectively.

The generator network $G$ is trained to confuse $D$ with visually plausible synthetic images, and the objective function could be written as,
\begin{equation}
\mathcal{L}_{adv_G} = - \: \mathbb{E}_{(I_y,\alpha)\sim P_y(I,\alpha)}\left[D(G(I_y,\alpha),\alpha)\right]
\end{equation}
Notably, since our model aims at rendering lifelike aging effects rather than transferring attributes of input face images, only correct attribute of young face images are fed into the generator in the training process.

\subsubsection{Identity Preservation Loss and Pixel-level Loss}
Although the goal of face aging is to modify a given face image to present aging effects, one key requirement is to preserve the identity-related information of the input.
To this end, we adopt the identity preserving loss to minimize the distance between input and output of the generator in the feature space embedding personalized characteristics.
Specifically, we employ a pre-trained LightCNN model~\cite{wu2018light}, denoted as $\phi_{id}$, as the feature extractor and fix the parameters during the training process.
To be concrete, the identity preserving loss is defined on the output of both the last pooling layer and the fully connected layer of $\phi_{id}$, which could be formulated as,
\begin{align}
\mathcal{L}_{id} & = \: \mathbb{E}_{(I_y,\alpha)\sim P_y(I,\alpha)}\left[\left\Vert \phi_{id}^{pool}(G(I_y,\alpha))-\phi_{id}^{pool}(I_y)\right\Vert_F^2\right]\nonumber\\
                 & + \: \mathbb{E}_{(I_y,\alpha)\sim P_y(I,\alpha)}\left[\left\Vert \phi_{id}^{fc}(G(I_y,\alpha))-\phi_{id}^{fc}(I_y)\right\Vert_2^2\right]
\end{align}
where $\phi_{id}^{pool}$ and $\phi_{id}^{fc}$ denote the output of the last pooling layer and the fully connected layer, respectively.
Additionally, a pixel-level loss is also adopted to maintain the consistency of low-level image content between the input and output of the generator, which could be written as,
\begin{equation}
\mathcal{L}_{pix} = \: \mathbb{E}_{(I_y,\alpha)\sim P_y(I,\alpha)}\left[\left\Vert G(I_y,\alpha_y)-I_y \right\Vert_2^2\right]
\end{equation}

\subsubsection{Overall objective}
To generate photo-realistic aged face with attributes faithful to the corresponding input, the the overall objective function for optimizing the discriminator $D$ could be formulated as
\begin{equation}
\min_{\Vert D\Vert_L\leq1} \: \mathcal{L}_{D} = \mathcal{L}_{adv_D} = \lambda_{att}\: \mathcal{L}_{adv\_att} + \mathcal{L}_{adv\_auth}
\end{equation}
where $\Vert D\Vert_L\leq1$ denotes the 1-Lipschitz constraint~\cite{Arjovsky2017WGAN} imposed on $D$, and is implemented by gradient penalty as proposed in WGAN-GP~\cite{gulrajani2017improved}.
The objective for the generator $G$ could be written as
\begin{equation}
\min_{G} \: \mathcal{L}_{G} = \mathcal{L}_{adv_G} + \lambda_{id}\mathcal{L}_{id} + \lambda_{pix}\mathcal{L}_{pix}
\end{equation}
where $\lambda_{id}$ and $\lambda_{pix}$ are hyperparameters for balancing the importance of $\mathcal{L}_{id}$ and $\mathcal{L}_{pix}$ w.r.t. the adversarial loss term, respectively.
$G$ and $D$ are trained alternatively until reaching the optimality.

\begin{figure*}[ht]
\begin{center}
\includegraphics[width=1.0\linewidth]{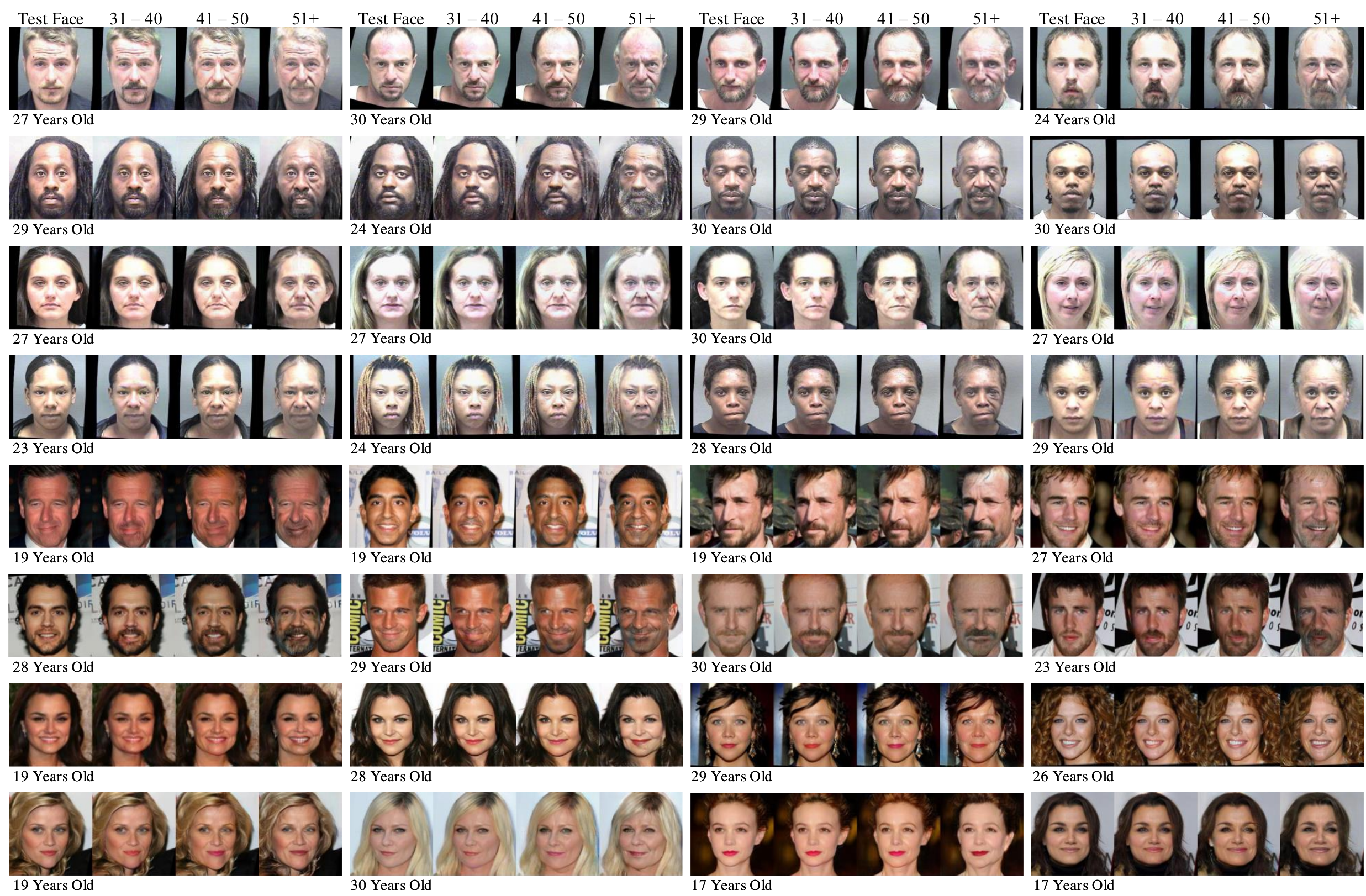}
\end{center}
\caption{Sample face aging results on MORPH (first four rows) and CACD (last four rows). The leftmost image of each result is the input test face image (age labeled below) and subsequent 3 images are synthesized elderly face images of the same subject in age group 31-40, 41-50 and 51+, respectively. Zoom in for a better view of aging details.}
\label{fig:sample_results_combined}
\end{figure*}

\section{Experiments}\label{sec:experiments}
Extensive experiments are conducted to validate the proposed \aaagan~in generating realistic and attribute-consistent aged face images.
In this section, we first introduce face aging datasets and then present implementation details of our model.
After that, extensive qualitative and quantitative results are reported to demonstrate the effectiveness of the proposed method.
Finally, ablation study is conducted to further explore the contribution of each component of our model.

\subsection{Face Aging Datasets}
Two publicly available face aging datasets, MORPH~\cite{ricanek2006morph} and CACD~\cite{chen2015face}, are employed in our experiments for both training and testing.
\textbf{MORPH} contains 55,134 face images of 13,000 people, covering an age span of 16 to 77. Face images in MORPH capture near-frontal faces of collaborative subjects under uniform and moderate illumination with simple background.
\textbf{CACD} contains 163,446 photos of 2,000 celebrities obtained in much less controlled (in-the-wild) conditions compared to MORPH. Consequently, large variations in terms of pose, illumination, and expression (PIE variations) exist in CACD. Besides, due to the fact that images in CACD are collected via Google Image Search, there are mismatches between faces and associated labels provided (i.e.~name and age), making it a very challenging dataset for accurate modeling of the face aging process.

Another two face datasets, FG-NET~\cite{FGNET} and CelebA~\cite{liu2015faceattributes}, are employed as test datasets to validate the generalization ability of the proposed model.
\textbf{FG-NET} contains 1,002 face portraits of 82 subjects and is widely adopted in the test phase of previous works~\cite{kemelmacher2014illumination,tang2017personalized,zhang2017age,yang2017learning}.
\textbf{CelebA} is a large-scale face dataset featuring diverse facial attributes, which contains 202,599 face images with 40 attribute annotations for each sample.
Similar to CACD, images in CelebA are also captured in the wild and cover large pose variations as well as background clutter.

\subsection{Implementation Details}
\textbf{Image Normalization.}
Following the convention of previous studies~\cite{wang2016recurrent,duong2017temporal,zhang2017age,yang2017learning,li2018global}, we adopt the age span of 10 years for each age group and only consider adult aging as both MORPH and CACD do not contain images of children.
In this way, faces are divided into four groups in terms of age, i.e.~30-, 31-40, 41-50, 51+, and only age translations from 30- to the other three age groups are considered.
All face images in MORPH and CACD are aligned according to eye locations detected using MTCNN~\cite{zhang2016joint} and then cropped into size $256\times 256\times 3$.
After face normalization, 51,822 and 163,355 face images from MORPH and CACD are used in the experiments, respectively.
As for FG-NET and CelebA, images are normalized according to facial landmarks provided along with the dataset.
In total, 984 faces from FG-NET and 10,000 images randomly sampled from CelebA are used as testing data in the cross-dataset validation experiment.

\textbf{Facial Attribute Labeling.} As for facial attribute labeling, MORPH provides researchers with labels including age, gender, and race for each image. We choose `gender' and `race' to be the attributes required to be preserved, since these two attributes are guaranteed to remain unchanged during the natural aging process, and are relatively objective compared to attributes such as `attractive' or `chubby' used in CelebA.
For CACD, we go through the name list of celebrities and label corresponding images accordingly. This introduces noise in attribute labels due to the mismatching between annotated names and actual faces presented, which further increases the difficulty for our method to achieve a satisfying performance on this dataset.
Since face images with race other than `white' only take a small portion of the entire dataset, we only select `gender' as the attribute to preserve.
Facial attributes of FG-NET are detected via public face analysis APIs of Face++\footnote{According to API documentations on the official website of Face++ (\url{https://console.faceplusplus.com}), the latest update of face APIs (Analyze API and Compare API) was in March, 2017. All quantitative results from Face++ API were obtained in August, 2019.}~\cite{face2018toolkit}, and for CelebA, we simply adopt facial attribute annotations provided along with the dataset.
It is worthwhile to note that the proposed model is highly expandable, as researchers may choose whatever attributes to preserve by simply incorporating them in the conditional facial attribute vector.

\textbf{Training Configurations.} We choose Adam to be the optimizer of both $G$ and $D$ with learning rate $1e^{-4}$.
As for trade-off parameters, $\lambda_{att}$, $\lambda_{pix}$ and $\lambda_{id}$ are set to $0.75$, $8.0$ and $0.02$, respectively.
The identity preserving loss is applied at every generator iteration, and the pixel-level loss is employed every 5 generator iterations, creating sufficient room for the generator to manipulate the input image.
On both MORPH and CACD, the model is trained with batch-size of 16 for 30 epochs, and all experiments are conducted under 5-fold cross-validation.

\subsection{Benchmark Methods}
To demonstrate the effectiveness of our model in performing lifelike and accurate age progression, six benchmark methods (CONGRE~\cite{suo2012concatenational}, HFA~\cite{yang2016face}, CAAE~\cite{zhang2017age}, GLCA-GAN~\cite{li2018global}, IPC-GAN~\cite{wang2018face_aging}, and PSD-GAN~\cite{yang2017learning}) are selected for comparison.

Specifically, CONGRE~\cite{suo2012concatenational} and HFA~\cite{yang2016face} do not adopt GAN-based framework thus only participate in the comparison of visual results for fairness, and results of other methods are considered as benchmarks for both qualitative and quantitative comparisons.
As for CAAE and IPC-GAN, code provided by corresponding authors~\footnote{CAAE: https://github.com/ZZUTK/Face-Aging-CAAE}~\footnote{IPC-GAN: https://github.com/dawei6875797/Face-Aging-with-Identity-Preserved-Conditional-Generative-Adversarial-Networks} are used for reproducing results, and hyper-parameters are fine-tuned to obtain the optimal results.
Since GLCA-GAN requires fine-grained cropping of facial components, original experimental results are obtained from the authors and directly used for evaluation.
As for PSD-GAN, we re-implemented the model and the VGG network in the discriminator is pre-trained on the same set of training samples as the entire GAN-based framework.

\begin{figure}[ht]
\begin{center}
\includegraphics[width=1.0\linewidth]{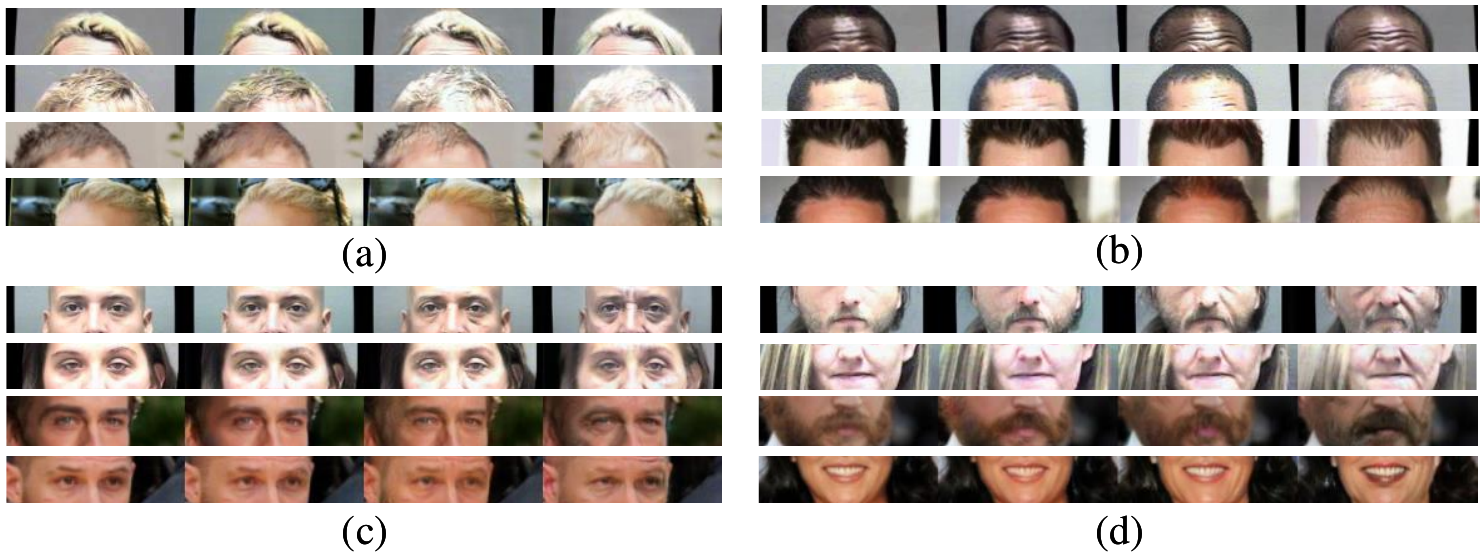}
\end{center}
\caption{Illustration of visual fidelity for different facial components: (a) Hair whitening; (b) Forehead wrinkle and receding hairline; (c) Eye region aging; (d) Mouth region aging. Zoom in for a better view of details.
}
\label{fig:aging_details}
\end{figure}

\begin{figure*}[ht]
\begin{center}
\includegraphics[width=0.95\linewidth]{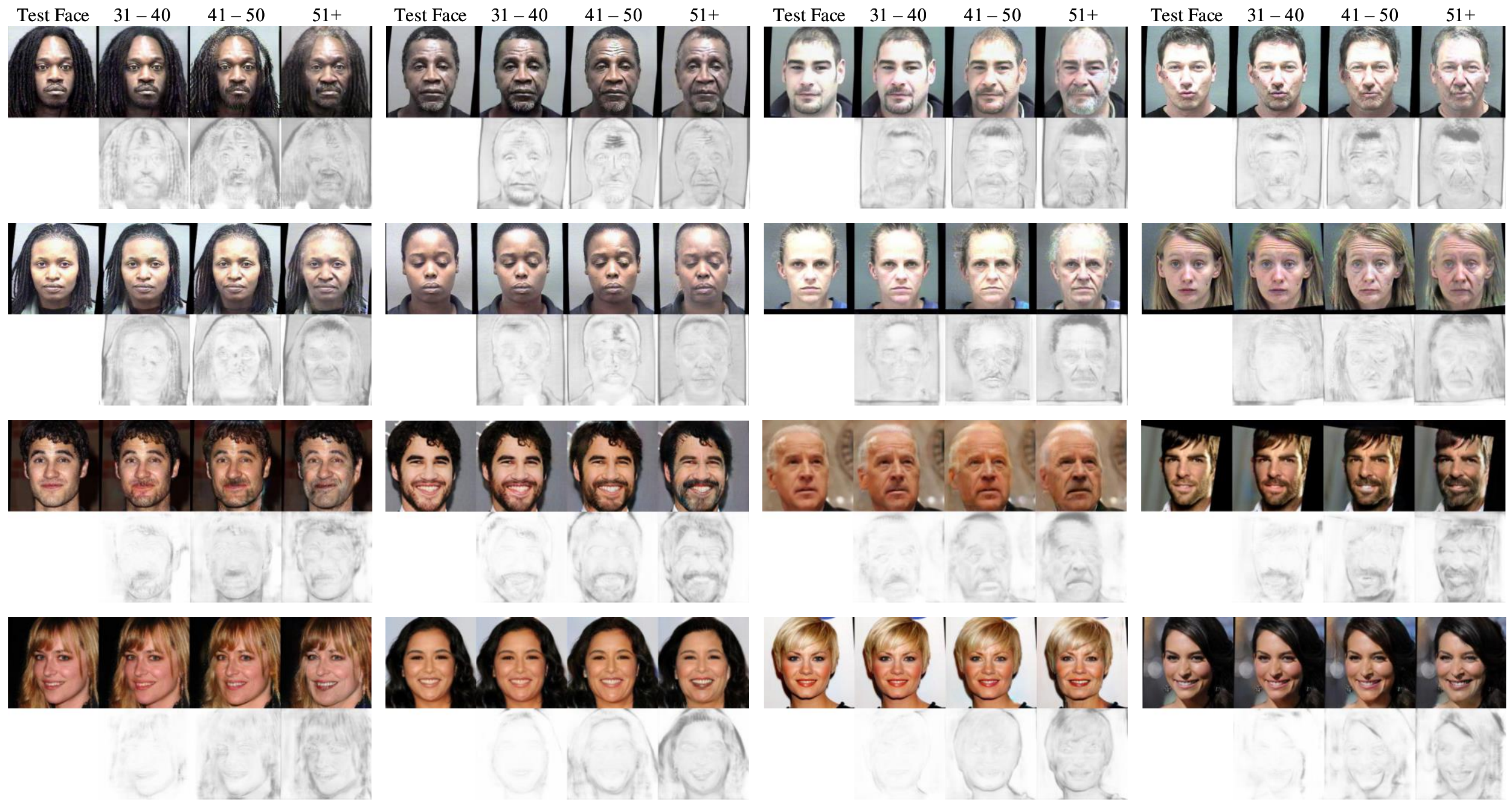}
\end{center}
\caption{Illustration of face aging results and corresponding attention maps on MORPH (first two rows) and CACD (second two rows). Darker regions suggest those areas of the face image receive more attention in the generation process, and brighter regions indicate that more information is retained from the original input image. Zoom in for a better view of aging details.}
\label{fig:attention_maps_Morph}
\end{figure*}

\subsection{Qualitative Evaluation of \aaagan}
\subsubsection{Face aging results on MORPH and CACD}\label{sec:sample_face_aging_results}
Sample face aging results on MORPH and CACD are shown in Fig.~\ref{fig:sample_results_combined}.
For each result, the leftmost image shows the test face under 30 years old, and the subsequent three images are synthesized aged faces in age group 31-40, 41-50 and 51+, respectively.
Although input face images cover a wide range of gender, race, pose, and expression, the proposed method could generate visually appealing face aging results with coherent and diverse signs of age progression, including bald forehead, white hair, laugh lines, etc.
Notably, compared to MORPH, generated faces in CACD present more fine-grained and subtler signs of aging, especially for female subjects.
This observation reflects the difference in data distributions of MORPH and CACD, as CACD mainly contains face images of celebrities with apparent make-up, making them look generally younger than faces in MORPH.
Quantitative results reported in Sec.~\ref{sec:aging_accuracy} also confirms our conlusion.
Closer inspect of aging details in different facial regions are presented in Fig.~\ref{fig:aging_details}.

\begin{figure*}[ht]
\begin{center}
\includegraphics[width=0.95\linewidth]{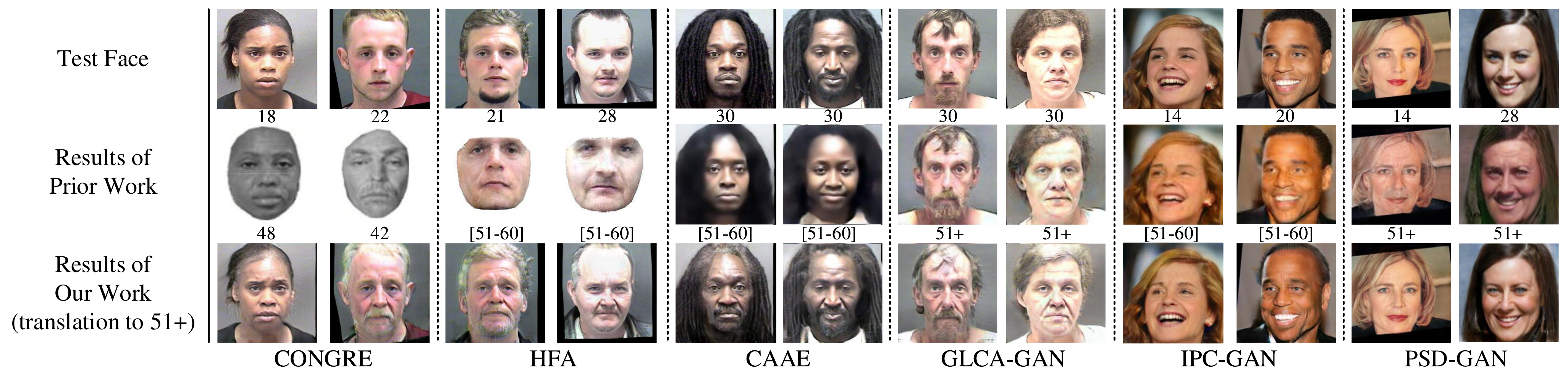}
\end{center}
\caption{Performance comparison with prior work on MORPH and CACD.
Samples results of six benchmark methods are presented in the second row with the target age (group) labeled below.
Test face images and results obtained by the proposed model are shown in the first and last row, respectively. Zoom in for a better comparison of aging details.}
\label{fig:visual_compare}
\end{figure*}

\begin{figure*}[ht]
\begin{center}
\includegraphics[width=0.9\linewidth]{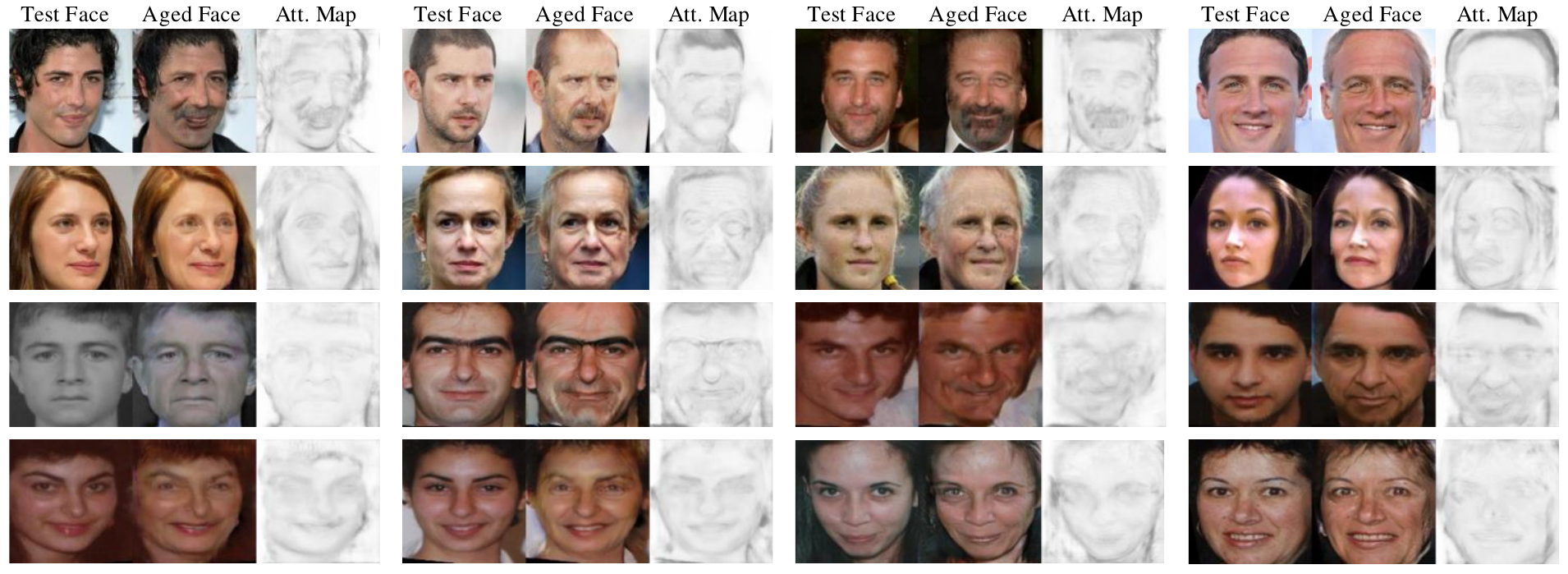}
\end{center}
\caption{Sample results achieved on CelebA (first two rows) and FG-NET (last two rows) with the model trained on CACD.
For each sample, the first image is the test face and the image in the middle shows the aging result. The attention map is (Att. Map) shown on the right for each result. Zoom in for a better view of aging details.}
\label{fig:inter_dataset}
\end{figure*}

\subsubsection{Demonstration of Spatial Attention Maps}
The attention mechanism is employed in our model to restrict image modifications within age-related regions.
Fig.~\ref{fig:attention_maps_Morph} shows sample face aging results with corresponding attention maps, indicating the contribution of input images to generated aged faces.
It could be observed that darker regions, which represent image areas receiving more attention in the generation process, are distributed mainly in facial areas closely related to signs of aging (e.g.~hair, forehead, and mouth).
Image content in these areas is modified to reflect age changes.
On the other hand, pixels located in brighter regions are mainly retained from the original input image.
This enables the generator to focus on synthesizing signs of aging, which is helpful for both preserving fine-grained textural details in the input image and improving the visual fidelity of generation results.

\subsubsection{Comparison with Prior Work}
To further demonstrate the effectiveness of our model, performance comparison is conducted between the proposed method and prior work on MORPH and CACD.
Six face aging methods, including both traditional (CONGRE~\cite{suo2012concatenational} and HFA~\cite{yang2016face}) and GAN-based models (CAAE~\cite{zhang2017age}, GLCA-GAN~\cite{li2018global}, IPC-GAN~\cite{wang2018face_aging}, and PSD-GAN~\cite{yang2017learning}), are considered as benchmarks, and comparison results are shown in Fig.~\ref{fig:visual_compare}.

Clearly, CONGRE and HFA only render subtle aging effects within the facial area while our method could also vividly simulate the process of hair whitening and hairline receding.
As for CAAE, due to its incapability in jointly modeling age progression and identity translation, over-smoothed faces are generated and signs of aging could hardly be observed.

Although more obvious aging effects could be seen in results of GLCA-GAN and IPC-GAN, they are originally designed for face images of size $128\times 128$ while our method works on higher resolution ($2\times$) with rich and enhanced details.
In addition, GLCA-GAN adopts local generator networks to emphasize aging patterns in facial patches of forehead, eyes, and mouth, but it overlooks the importance of the hair region which is also critical in reflecting age changes.

With the aid of multi-level face representations extracted by the pre-trained deep network in the discriminator, PSD-GAN is able to generate aged faces with high visual fidelity.
However, they suffer from obvious color distortions (hair and background) since the value of every single pixel is re-estimated by the generator rather than retained from the input image.
Moreover, due to the lack of prior knowledge of input faces, masculine facial characteristics (e.g.~stubble above the mouth) emerge in the aged faces, making the generation results look much less natural.

\subsubsection{Cross-dataset Validation}
To evaluate the generalization ability of our model, cross-dataset validation experiments are conducted on FG-NET and CelebA with the model trained on CACD and results are shown in Fig.~\ref{fig:inter_dataset}.
Although input faces are sampled from data distributions different from the training set, visually plausible aging results could still be obtained via the proposed method, demonstrating its effectiveness in dealing with unseen face images.
Notably, although test faces follow different data distributions, activated regions in the attention map still concentrate on facial areas closely related to aging effects (e.g.~white hair and laugh lines), which helps improve the visual quality of generation results by restricting the image region being modified.

\subsection{Quantitative Evaluation}\label{sec:quantitative_results}
Apart from visual fidelity, the performance of the proposed model could also be quantitatively evaluated in the following aspects:
\begin{itemize}
\item\textbf{Aging Accuracy:} Synthesized aged faces are expected to present accurate aging signs that make them fall into the target age group.
\item\textbf{Identity Preservation:} Personal characteristics of input faces are supposed to be preserved in generation results.
\item\textbf{Attribute Consistency:} Besides identity, facial attributes that should remain stable in the natural aging process, such as gender and race, are expected to be consistent between input and generated faces.
\end{itemize}

To evaluate the performance of the proposed method objectively, measurements of all metrics are conducted via the public face analysis API of Face++~\cite{face2018toolkit}.
To unify the evaluation criteria, all metrics are computed based on results obtained by the Face++ API, rather than annotations provided along with the dataset.
For each evaluation metric, the results of four other GAN-based frameworks (CAAE, GLCA-GAN, IPC-GAN, and PSD-GAN) are also reported for comparison.

\begin{table*}[!t]
\caption{Results of age estimation on MORPH and CACD.}
\label{table:AgeAcc}
\centering
\ra{1.1}
\begin{tabular} {@{}l lll c l lll@{}}
\toprule
          \multicolumn{4}{c}{MORPH}                           &\phantom{a} & \multicolumn{4}{c}{CACD}                 \\
          \cmidrule{1-4}                                                    \cmidrule{6-9}
Age group & \multicolumn{1}{c}{31 - 40} & \multicolumn{1}{c}{41 - 50} & \multicolumn{1}{c}{51 +} &\phantom{a} &
Age group & \multicolumn{1}{c}{31 - 40} & \multicolumn{1}{c}{41 - 50} & \multicolumn{1}{c}{51 +}  \\
          \cmidrule{1-4}                                                    \cmidrule{6-9}
Generic   & $38.60\pm7.43$ & $47.74\pm8.30$ & $57.25\pm8.29$  &\phantom{a} & Generic   & $38.50\pm9.66$ & $46.53\pm10.69$ & $53.41\pm12.41$ \\
\midrule
          \multicolumn{4}{c}{Distributions of estimated ages} &\phantom{a} & \multicolumn{4}{c}{Distributions of estimated ages} \\
          \cmidrule{1-4}                                                    \cmidrule{6-9}
CAAE      & $28.52\pm5.26$ & $32.25\pm6.40$ & $35.83\pm7.49$  &\phantom{a} & CAAE      & $32.84\pm7.67$  & $36.21\pm8.56$  & $39.35\pm9.65$  \\
GLCA-GAN  & $43.79\pm6.21$ & $48.33\pm6.85$ & $53.25\pm7.71$  &\phantom{a} & GLCA-GAN  & $38.15\pm8.61$  & $45.53\pm9.04$  & $53.30\pm9.76$  \\
IPC-GAN   & $36.45\pm5.62$ & $45.87\pm5.80$ & $55.63\pm5.65$  &\phantom{a} & IPC-GAN   & $38.04\pm8.59$  & $47.75\pm7.38$  & $55.49\pm8.71$ \\
PSD-GAN   & $39.80\pm6.90$ & $50.09\pm6.75$ & $58.36\pm6.91$  &\phantom{a} & PSD-GAN   & $42.97\pm9.80$  & $49.53\pm9.23$  & $55.75\pm10.13$ \\
Ours      & $38.84\pm7.42$ & $47.84\pm7.03$ & $56.68\pm6.78$  &\phantom{a} & Ours      & $38.26\pm9.36$  & $47.25\pm9.73$  & $54.05\pm9.17$ \\
\\
\multicolumn{4}{c}{Difference of mean ages (against generic faces)} &\phantom{a} & \multicolumn{4}{c}{Difference of mean ages (against generic faces)} \\
          \cmidrule{1-4}                                                    \cmidrule{6-9}
CAAE      & $-10.08$       & $-15.49$       & $-21.42$        &\phantom{a} & CAAE      & $-5.66$         & $-10.32$        & $-14.06$  \\
GLCA-GAN  & $+5.19 $       & $+0.59 $       & $-4.00 $        &\phantom{a} & GLCA-GAN  & $-0.35$         & $-1.00 $        & $-0.11 $  \\
IPC-GAN   & $-2.15 $       & $-1.87 $       & $-1.62 $        &\phantom{a} & IPC-GAN   & $-0.46$         & $+1.22 $        & $+2.08 $  \\
PSD-GAN   & $+1.20 $       & $+2.35 $       & $+1.11 $        &\phantom{a} & PSD-GAN   & $+4.47$         & $+3.00 $        & $+2.34 $  \\
Ours      & $+0.24 $       & $+0.10 $       & $-0.57 $        &\phantom{a} & Ours      & $-0.24$         & $+0.72 $        & $+0.64 $  \\
\bottomrule
\end{tabular}
\end{table*}

\subsubsection{Aging Accuracy}\label{sec:aging_accuracy}
The goal of face aging is rendering a given face with aging effects to predict its appearance in the future.
Therefore, the age distribution of generated aged faces should match that of real faces from the same age group to represent accurate age simulation.
In this experiment, age distributions of both generic and synthetic faces are estimated and compared for all three target age groups (31-40, 41-50, 51+).
Results of age estimation on MORPH and CACD are shown in TABLE~\ref{table:AgeAcc}.
Clearly, mean estimated ages of synthesized faces show the trend of increasing age (38.84, 47.84, 56.68 on MORPH, and 38.26, 47.25, 54.05 on CACD), and are close to that of real faces (38.60, 47.74, 57.25 on MORPH, and 38.50, 46.53, 53.41 on CACD), demonstrating the effectiveness of our model in accurately simulating age progression with various time intervals.

\begin{figure}[!t]
\begin{center}
\includegraphics[width=1.0\linewidth]{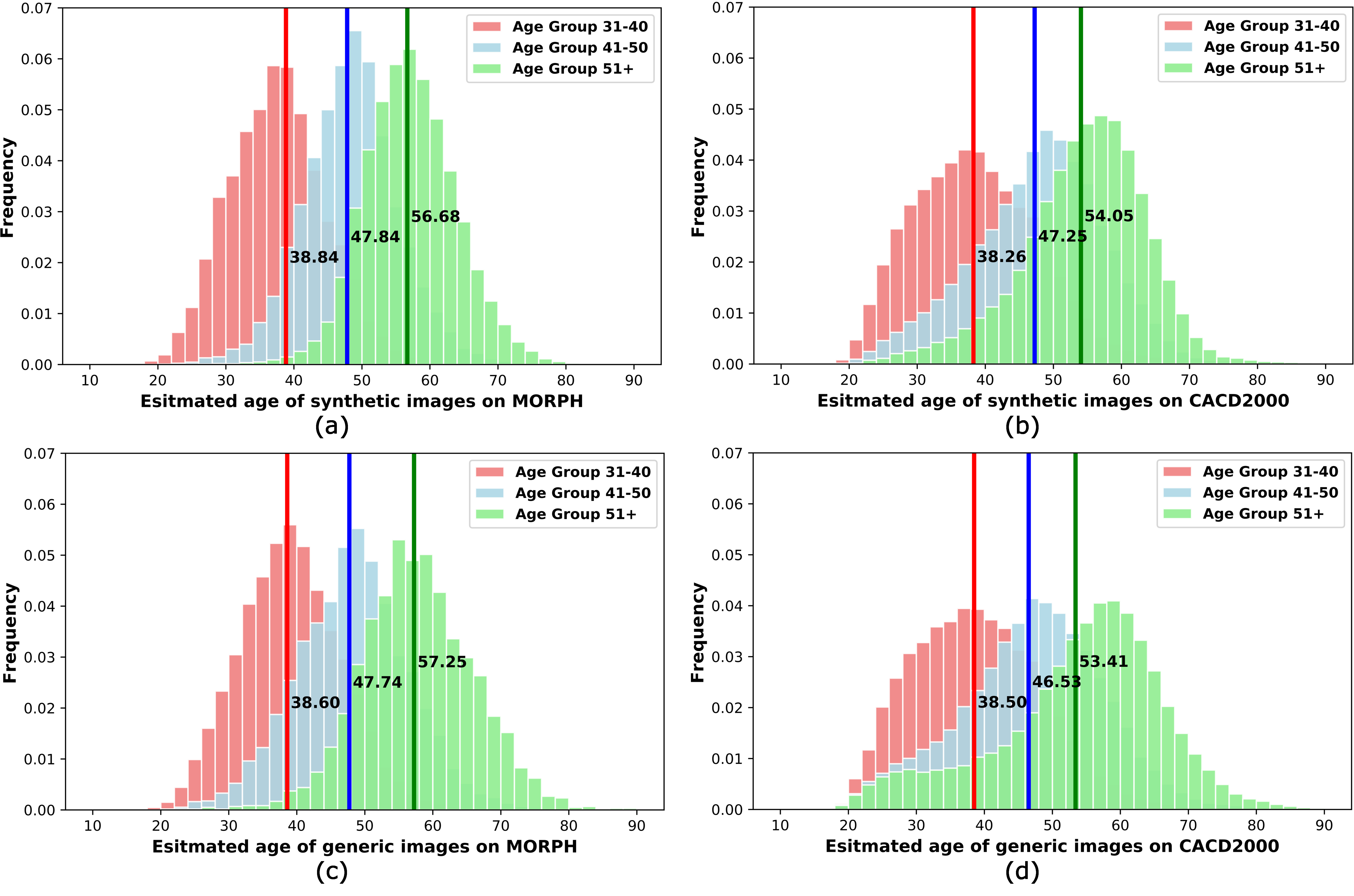}
\end{center}
\caption{Estimated age distributions of (a) synthetic faces on MORPH; (b) synthetic faces on CACD; (c) generic faces on MORPH;
(d) generic faces on CACD.}
\label{fig:histograms}
\end{figure}

Compared to our method, CAAE produces over-smoothed face images with subtle changes of appearance on both datasets, leading to insufficient facial changes and large errors in estimated ages.
Although much more obvious aging signs could be synthesized by GLCA-GAN, on MORPH, stitching outputs of several local aging networks introduce additional ghosting artifacts in generation results of age group 31-40, causing large errors ($+5.19$) in estimated ages.
With the aid of a pre-trained VGG network adopted in the discriminator, PSD-GAN could effectively extract multi-level age-related representations and generate faces with clear aging signs.
However, due to the matching ambiguity of facial attributes, translations in gender (female to male) take place when synthesizing aged faces, causing the overall age of generation results to be higher than generic faces, especially on CACD.
This observation could be confirmed by the visual comparison between our method and PSD-GAN shown in Fig.~\ref{fig:visual_compare} as well as preservation rate of `gender' on CACD reported in TABLE~\ref{table:AttPreserve}.

Further comparisons between detailed age distributions between real and generated face images are shown in Fig.~\ref{fig:histograms}.
For each age group, it could be observed that the age distribution on MORPH is more concentrated than that on CACD by comparing Fig.~\ref{fig:histograms} (c) and Fig.~\ref{fig:histograms} (d).
This is because faces in CACD are obtained online via image search engine thus contain noisy age labels.
By comparing Fig.~\ref{fig:histograms} (a) and Fig.~\ref{fig:histograms} (c) as well as Fig.~\ref{fig:histograms} (b) and Fig.~\ref{fig:histograms} (d), it is clear that age distributions of faces generated by our model well match that of real faces, indicating the effectiveness of the proposed method in rendering accurate age translations.

\begin{table*}[!t]
\caption{Resutls on face verification on MORPH and CACD.}
\label{table:IdPreserve}
\centering
\ra{1.1}
\begin{tabular} {@{}l lll c l lll@{}}
\toprule
          \multicolumn{4}{c}{MORPH}                           &\phantom{a} & \multicolumn{4}{c}{CACD}                 \\
          \cmidrule{1-4}                                                    \cmidrule{6-9}
Age group & \multicolumn{1}{c}{31 - 40} & \multicolumn{1}{c}{41 - 50} & \multicolumn{1}{c}{51 +} &\phantom{a} &
Age group & \multicolumn{1}{c}{31 - 40} & \multicolumn{1}{c}{41 - 50} & \multicolumn{1}{c}{51 +}  \\
\midrule
           \multicolumn{4}{c}{Face Verification Confidence}    &\phantom{a} &           \multicolumn{4}{c}{Face Verification Confidence} \\
           \cmidrule{1-4}                                                               \cmidrule{6-9}
CAAE     & $66.55\pm9.99$  & $65.20\pm10.06$ & $63.32\pm10.17$ &\phantom{a} & CAAE     & $60.67\pm10.65$ & $59.30\pm10.73$ & $57.98\pm10.82$ \\
GLCA-GAN & $91.84\pm2.33$  & $90.42\pm2.82$  & $86.89\pm4.20$  &\phantom{a} & GLCA-GAN & $86.80\pm5.65$  & $85.30\pm5.48$  & $84.69\pm5.47$  \\
IPC-GAN  & $95.69\pm0.45$  & $93.92\pm1.01$  & $90.73\pm2.35$  &\phantom{a} & IPC-GAN  & $91.86\pm2.18$  & $86.36\pm4.86$  & $87.85\pm4.25$  \\
PSD-GAN  & $95.48\pm0.80$  & $92.64\pm1.99$  & $89.00\pm3.29$  &\phantom{a} & PSD-GAN  & $94.82\pm2.58$  & $90.14\pm5.53$  & $90.38\pm4.78$  \\
Ours     & $95.92\pm0.66$  & $92.76\pm2.59$  & $88.81\pm3.92$  &\phantom{a} & Ours     & $96.19\pm1.77$  & $94.39\pm3.23$  & $90.69\pm4.56$  \\
\\
           \multicolumn{4}{c}{Face Verification Rate (\%)}     &\phantom{a} &          \multicolumn{4}{c}{Face Verification Rate (\%)} \\
           \cmidrule{1-4}                                                              \cmidrule{6-9}
CAAE     & 24.28           & 20.05           & 14.42           &\phantom{a} & CAAE     & 9.20            & 7.04            & 5.10            \\
GLCA-GAN & 100.00          & 99.97           & 98.99           &\phantom{a} & GLCA-GAN & 96.09           & 95.79           & 95.29           \\
IPC-GAN  & 100.00          & 100.00          & 99.48           &\phantom{a} & IPC-GAN  & 100.00          & 97.95           & 97.36           \\
PSD-GAN  & 100.00          & 100.00          & 99.42           &\phantom{a} & PSD-GAN  & 99.83           & 97.67           & 98.50           \\
Ours     & 100.00          & 100.00          & 99.53           &\phantom{a} & Ours     & 99.94           & 99.61           & 98.85           \\
\bottomrule
\end{tabular}
\end{table*}

\begin{table*}[!t]
\caption{Preservation rate of facial attributes on MORPH and CACD.}
\label{table:AttPreserve}
\centering
\begin{tabular} {@{}l lll r lll r lll@{}}
\toprule
           & \multicolumn{7}{c}{MORPH}       &\phantom{a} & \multicolumn{3}{c}{CACD} \\
             \cmidrule{2-8}                                 \cmidrule{10-12}
           & \multicolumn{3}{c}{Gender (\%)} &\phantom{a} & \multicolumn{3}{c}{Race (\%)} &\phantom{a} & \multicolumn{3}{c}{Gender (\%)} \\
             \cmidrule{2-4}                                 \cmidrule{6-8}                               \cmidrule{10-12}
Age group  & 31 - 40 & 41 - 50 & 51 +        &\phantom{a} & 31 - 40 & 41 - 50 & 51 +      &\phantom{a} & 31 - 40 & 41 - 50 & 51 +  \\
\midrule
CAAE       & 51.38   & 47.07   & 54.24       &\phantom{a} & 95.45   & 95.23   & 92.37     &\phantom{a} & 87.43   & 86.53   & 85.25 \\
GLCA-GAN   & 96.44   & 95.90   & 94.85       &\phantom{a} & 93.69   & 91.79   & 91.48     &\phantom{a} & 95.46   & 95.51   & 94.65 \\
IPC-GAN    & 96.87   & 97.45   & 96.75       &\phantom{a} & 97.11   & 96.88   & 90.57     &\phantom{a} & 94.79   & 90.18   & 93.24 \\
PSD-GAN    & 96.62   & 95.94   & 93.28       &\phantom{a} & 96.61   & 91.77   & 91.42     &\phantom{a} & 87.56   & 83.19   & 75.72 \\
Ours       & 97.41   & 97.58   & 96.92       &\phantom{a} & 97.68   & 96.36   & 93.28     &\phantom{a} & 99.00   & 98.59   & 98.00 \\
\bottomrule
\end{tabular}
\end{table*}

\subsubsection{Identity Preservation}
Besides rendering representative signs of aging, a face aging model is also expected to preserve personalized characteristics embedded in the input young face when synthesizing the corresponding aged face.
To this end, face verification experiments are carried out to measure the similarity between real faces from age group 30- and their age-progressed counterparts in age group 31-40, 41-50, and 51+, respectively.

Results of face verification, including confidence scores and verification rates (threshold set to 73.395@FAR=1e-5 for all experiments), are reported in TABLE~\ref{table:IdPreserve}.
On MORPH, our model achieves verification rates of $100.00\%$, $100.00\%$, and $99.53\%$ on translation to age group 31-40, 41-50, and 51+, respectively.
Although there are larger variations in pose, expression, and background textures in images of CACD, face verification rates of $99.94\%$, $99.61\%$, and $98.85\%$ are obtained on three age groups, demonstrating the effectiveness of the proposed method in preserving identity information.
Notably, as the time interval of age progression increases, both confidence scores and verification rates gradually decrease.
This is reasonable since a larger age gap is reflected in more obvious aging signs (e.g.~deeper wrinkles and eye bags), which may lower the similarity between faces from different age groups.

As for benchmark methods, personalized facial features are failed to be preserved in heavily blurred aging results generated by CAAE, causing poor face verification performance.
To preserve as many facial details in the input face image as much, a residual connection is adopted in GLCA-GAN by adding the input face image to the output of the generator. However, separate translations of different facial components inevitably introduce extra distortions and ghosting artifacts, which lead to a slightly lower verification rate on generated face images of 51+.
The performance of PSD-GAN on identity preservation is very close to ours, and the verification rate between 30- and 51+ on MORPH ($99.42\%$) is clearly higher than reported in~\cite{yang2017learning} ($93.09\%$), indicating the quality of our re-implementation.

\subsubsection{Facial Attribute Preservation}
In this experiment, the performance of facial attribute preservation is evaluated by comparing attributes of generic and synthetic faces estimated by the Face++ API, and results are shown in TABLE~\ref{table:AttPreserve}.
On MORPH, our model achieves preservation rate of $97.41\%$, $97.58\%$, $96.92\%$ on `Gender' and $97.68\%$, $96.36\%$, $93.28\%$ on `Race', for age mappings from 30- to 31-40, 41-50, and 51+, respectively.
As for CACD, $99.00\%$, $98.59\%$, and $98.00\%$ of generated faces in age group 31-40, 41-50, and 51+ have facial attributes consistent with the corresponding input, respectively.
Notably, similar to results of face verification experiments, with the age gap increases, preservation rates of attributes decrease due to larger variations of facial appearance.


According to TABLE~\ref{table:AttPreserve}, it could be observed that the proposed method consistently outperforms other benchmarks by a clear margin in all cases, demonstrating the effectiveness of our model in preserving facial attributes beyond identity information during the face aging process.
Specifically, gender characteristics are lost along with identity information in faces generated by CAAE, causing large errors in preservation rate on `Gender'.
Among all four benchmarks, GLCA-GAN and IPC-GAN give better performance on maintaining facial attribute consistency.
This is because they are applied to face images of lower resolution ($128\times 128$) with less textural details, which reduces the chance of introducing distortions of fine-grained image content.
Although PSD-GAN achieves good results on aging accuracy and face verification, it suffers from inconsistent facial attributes between input and generated faces, due to the lack of prior knowledge regarding the input image.

\begin{table*}[!t]
\caption{Performance comparison on aging accuracy between variants of the proposed model (difference of mean ages against generic faces are shown in brackets).}
\label{table:AgeAccAblation}
\centering
\ra{1.1}
\begin{tabular}{@{}l lll @{}}
\toprule
Age group & \multicolumn{1}{c}{31-40} & \multicolumn{1}{c}{41-50} & \multicolumn{1}{c}{51+}    \\
\midrule
Generic   & $38.60\pm7.43$ & $47.74\pm8.30$ & $57.25\pm8.29$ \\
\midrule
Baseline  & $36.67\pm7.62\:(-2.17)$   & $45.21\pm7.98\:(-2.53)$  & $55.47\pm8.75\:(-1.78)$  \\
w/o FAE   & $37.32\pm7.56\:(-1.28)$   & $46.88\pm7.68\:(-0.86)$  & $55.53\pm8.31\:(-1.72)$  \\
w/o WMD   & $36.22\pm6.48\:(-2.38)$   & $46.47\pm7.83\:(-1.27)$  & $53.53\pm8.30\:(-3.72)$  \\
w/o AM  & $37.23\pm7.12\:(-1.37)$   & $48.18\pm7.55\:(+0.44)$  & $54.84\pm8.03\:(-2.41)$  \\
Proposed  & $38.84\pm7.42\:(+0.24)$   & $47.84\pm7.03\:(+0.10)$  & $56.68\pm6.78\:(-0.57)$  \\
\bottomrule
\end{tabular}
\end{table*}

\begin{table*}[!t]
\caption{Performance comparison on facial attribute preservation and face verification between variants of the proposed model..}
\label{table:AttPreIdPreAblation}
\centering
\ra{1.1}
\begin{tabular}{@{}l ccc | ccc | ccc | ccc@{}}
\toprule
          & \multicolumn{3}{c}{Gender Pre. Rate (\%)}  & \multicolumn{3}{c}{Race Pre. Rate (\%)} &  \multicolumn{3}{c}{Face Veri. Rate (\%)}   & \multicolumn{3}{c}{Face Veri. Score} \\
\midrule
Age group & 31-40 & 41-50 & 51+       & 31-40 & 41-50 & 51+       & 31-40 & 41-50   & 51+       & 31-40 & 41-50 & 51+ \\
\midrule
Baseline  & 97.05 & 95.35 & 92.20     & 97.04 & 94.85 & 91.18     & 100.00 & 99.99  & 97.66     & $95.75\pm0.72$ & $93.09\pm2.31$ & $86.23\pm4.77$ \\
w/o FAE   & 96.95 & 96.11 & 92.93     & 95.20 & 95.84 & 88.32     & 100.00 & 99.99  & 97.68     & $95.31\pm1.29$ & $93.26\pm2.20$ & $86.56\pm5.20$ \\
w/o WMD   & 96.84 & 96.93 & 96.14     & 97.44 & 96.74 & 91.66     & 100.00 & 99.96  & 97.02     & $95.68\pm1.03$ & $93.24\pm2.34$ & $88.41\pm3.99$ \\
w/o AM  & 96.90 & 96.27 & 94.95     & 97.69 & 95.89 & 91.57     & 100.00 & 99.93  & 98.44     & $95.65\pm0.90$ & $92.17\pm2.57$ & $86.31\pm4.47$ \\
Proposed  & 97.41 & 97.58 & 96.92     & 97.68 & 96.36 & 93.28     & 100.00 & 100.00 & 99.63     & $95.92\pm0.66$ & $92.76\pm2.59$ & $88.81\pm3.92$ \\
\bottomrule
\end{tabular}
\end{table*}

\subsection{Ablation Study}
In this subsection, experiments are carried out to comprehensively analyze the contribution of each component of the proposed model, namely, facial attribute embedding (FAE), wavelet-based multi-pathway discriminator (WMD), and attention mechanism (AM).
Specifically, `w/o FAE' denotes the setting that no facial attribute is considered as the conditional information, and both the generator and discriminator only receive image data as input.
For `w/o WMD', the proposed wavelet-based multi-pathway discriminator is replaced with an ordinary PatchGAN discrminator~\cite{isola2017image}.

\begin{figure}[ht]
\begin{center}
\includegraphics[width=1.0\linewidth]{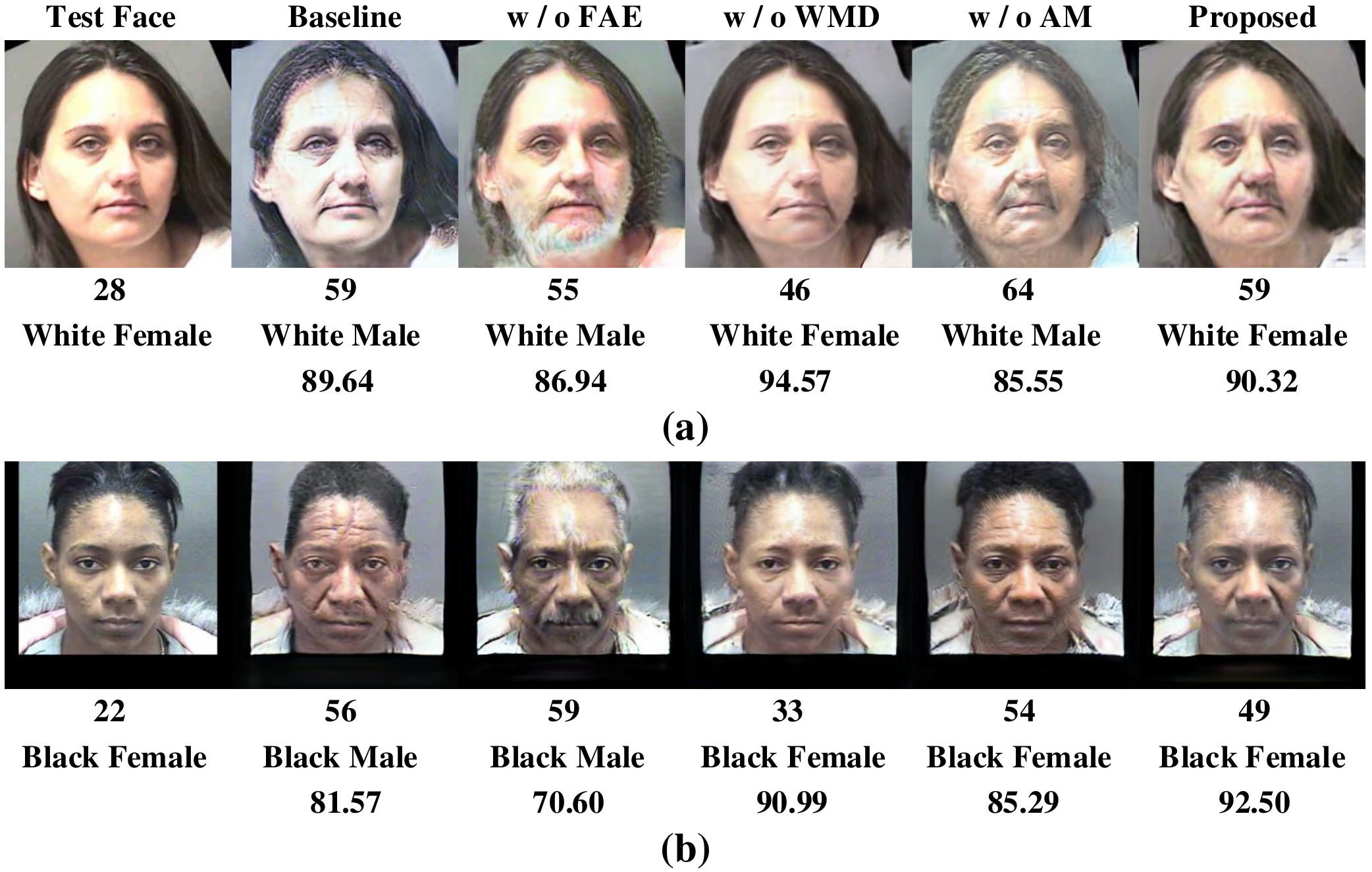}
\end{center}
\caption{Illustration of face aging results generated by different variants of the proposed model. For each subject, estimated ages (first row), faical attributes (second row), and confidence scores of face verification (third row) are reported. All quantitative results are obtained using Face++ API.}
\label{fig:ablation_study}
\end{figure}

Moreover, `w/o AM' refers to the variant without attention mechanism, where the generator works on synthesizing pixel at each location of the entire output image.
The impact of excluding each of these factors is studied in terms of visual fidelity, aging accuracy, identity verification, and facial attribute preservation.

Generation results obtained by different variants of the proposed model are shown in Fig.~\ref{fig:ablation_study}.
It could be observed that aged faces synthesized via the baseline model have severe ghosting artifacts (e.g.~hair area) and color distortion (e.g.~facial skin).
Since no semantic prior knowledge of the input face is considered in the aging process, masculine facial characteristics emerge and gender reversal takes place.
Similarly, results obtained under the setting `w/o FAE' also suffer from unnatural translations of facial attributes due to the lack of conditional information.
However, it could be noticed that involving WMD and AM helps to capture more representative age-related facial features (e.g.~beard in Fig.~\ref{fig:ablation_study} (a), white hair and mustache in Fig.~\ref{fig:ablation_study} (b)) and preserving image content in the input, respectively.
Notably, although aged faces generated under setting `baseline' and `w/o FAE' have limitations in maintaining the consistency of facial attributes, this is not clearly reflected in results of age estimation or face verification shown in TABLE~\ref{table:AgeAccAblation} and TABLE~\ref{table:AttPreIdPreAblation}.
Therefore, it could be concluded that merely enforcing identity consistency is insufficient in synthesizing aged faces reasonable in terms of facial attributes.

After adopting FAE, the unnatural translation of gender characteristics is greatly suppressed, as could be observed from results under `w/o WMD' and `w/o AM' in Fig.~\ref{fig:ablation_study}.
However, closer inspect would clearly reveal that replacing WMD with ordinary PatchGAN discriminator (`w/o WMD') damages the performance of the model in capturing age-related texture details, resulting in relatively larger errors in estimated ages.
As for results obtained under `w/o AM', distortions of color and image content (e.g.~facial contour and textual details of the hair area in Fig.~\ref{fig:ablation_study} (b)) as well as ghosting artifacts (e.g.~mouth and hair region in Fig.~\ref{fig:ablation_study} (a)) could be observed, leading to lower verification confidence and even incorrect facial attribute recognition results.
This confirms the contribution of the attention mechanism, that is, improving the visual fidelity of generation results by only attending to specific image regions closely related to age progression, and retaining textual details from the input face for the rest image areas.

Quantitative results for ablation study are reported in TABLE~\ref{table:AgeAccAblation} and TABLE~\ref{table:AttPreIdPreAblation}.
According to results in TABLE~\ref{table:AttPreIdPreAblation}, removing the facial attribute embedding component (`w/o FAE' and `Baseline') would cause obvious performance drop in preservation rate for both `gender' and `race'.
From another perspective of view, while face verification rates have already reached a high level (over $98.00\%$ for most cases), there is still relatively large room for improvement on preservation rates of facial attributes.
Therefore, we could conclude that facial attribute consistency is complementary to identity permanence in rendering natural and reasonable face aging results with high visual fidelity.

In addition, from results in TABLE~\ref{table:AgeAccAblation}, it could be concluded that adopting wavelet-based multi-pathway discriminator (WMD) reduces the gap between age distributions of real and synthesized faces for all age mappings.
This demonstrates the ability of WMD in capturing discriminative representations of age progression which is helpful in rendering more accurate signs of aging.
Moreover, introducing the attention mechanism helps comprehensively improve the performance of the model on all experiments, indicating its effectiveness in generating aged faces with high visual quality.

\section{Conclusion and Future Work}\label{sec:conclusion}
In this paper, an attribute-aware attentive face aging model, named as \aaagan, is proposed to overcome two major limitations of existing face aging methods, i.e.~unnatural translations of facial attributes and modifications to image contents irrelevant to age progression.
Specifically, facial attributes of input images are considered as conditional information and embedded to both the generator and discriminator to encourage attribute consistency.
Besides, the attention mechanism is adopted to restrict modifications to age-related regions and preserve image details from the input image for the rest area, improving the visual fidelity of generation results.
Moreover, a wavelet packet transform module is employed to extract textural features, and a multi-pathway discriminator is designed to capture age-related representations in multiple scales.
Extensive experimental results on MORPH and CACD demonstrate the effectiveness of the proposed model in rendering accurate aging effects while maintaining identity permanence and facial attribute consistency.

Although the proposed method achieves state-of-the-art performance in various experiments, it indeed has some limitations.
Since existing face aging datasets are heavily biased towards White and Black people, aging patterns of other ethnic groups (e.g.~Asian and Hispanic) receive much less attention.
Besides, child aging is not investigated in this work due to insufficient data of child faces at different ages.
Considering the above issues, collecting large-scale high-quality face images covering various ethnic and age groups could be one working direction in the future.

%
%

\ifCLASSOPTIONcaptionsoff
  \newpage
\fi



%


\bibliographystyle{IEEEtran}
\bibliography{IEEEabrv,IEEEbib}

%

\begin{IEEEbiography}[{\includegraphics[width=1in,height=1.25in,clip,keepaspectratio]{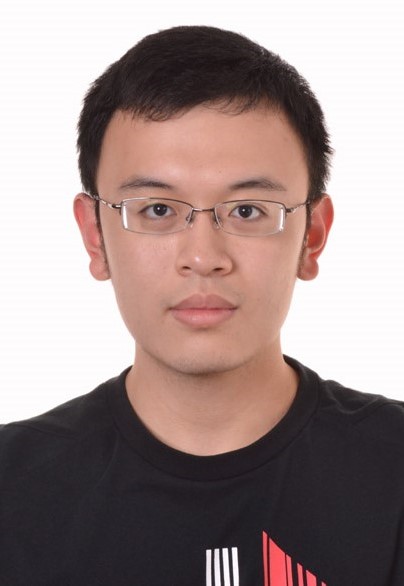}}]{Yunfan Liu}
received the B.E. degree in electronic engineering from Tsinghua University, Beijing, China, in 2015, the M.S. degree in electronic engineering: systems from University of Michigan, Ann Arbor, United States, in 2017.
He is currently pursuing the Ph.D. degree with the National Laboratory of Pattern Recognition, Center for Research on Intelligent Perception and Computing, Institute of Automation, Chinese Academy of Sciences, Beijing, China. His research interests include computer vision, pattern recognition, and machine learning.
\end{IEEEbiography}

\begin{IEEEbiography}[{\includegraphics[width=1in,height=1.25in,clip,keepaspectratio]{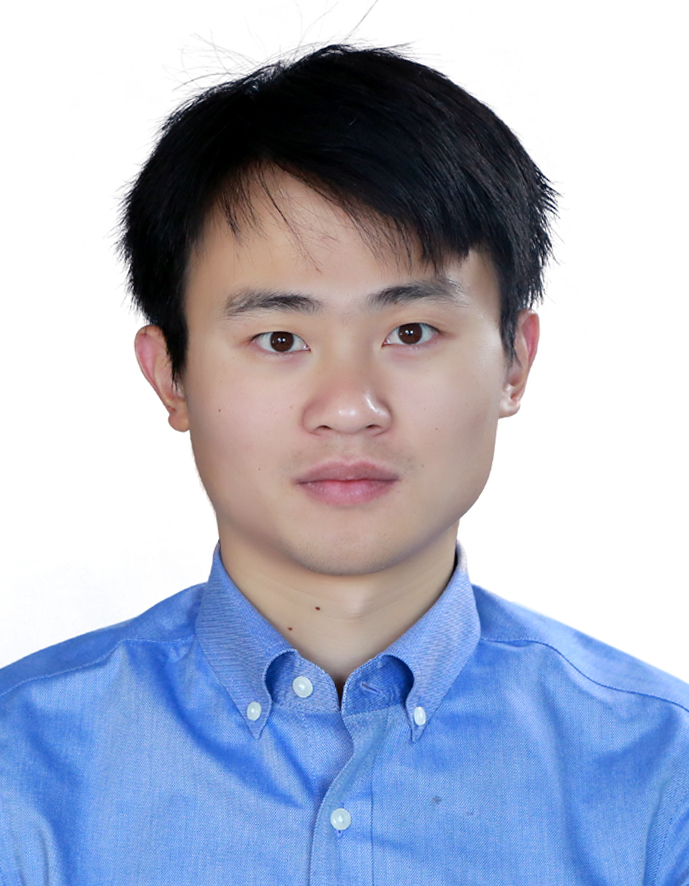}}]{Qi Li}
received the B.E. degree in automation from China University of Petroleum, Qingdao, China, in 2011, the Ph.D. degree in pattern recognition and intelligent systems from National Laboratory of Pattern Recognition, CASIA, Beijing, China, in 2016.
He is currently an assistant professor in the Center for Research on Intelligent Perception and Computing, National Laboratory of Pattern Recognition, CASIA.
His research interests include face recognition, computer vision and machine learning.
\end{IEEEbiography}

\vfill

\begin{IEEEbiography}[{\includegraphics[width=1in,height=1.25in,clip,keepaspectratio]{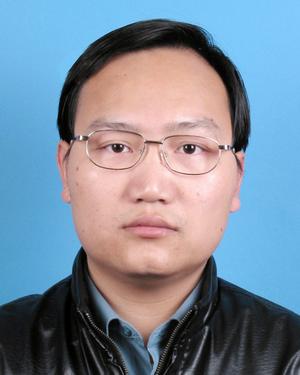}}]{Zhenan Sun}
received the B.S. degree in industrial automation from the Dalian University of Technology, Dalian, China, in 1999, the M.S. degree in system engineering from the Huazhong University of Science and Technology, Wuhan, China, in 2002, and the Ph.D. degree in pattern recognition and intelligent systems from the Chinese Academy of Sciences, Beijing, China, in 2006. Since 2006, he has
been a Faculty Member with the National Laboratory of Pattern Recognition, Institute of Automation, Chinese Academy of Sciences, where he is currently a Professor. He has authored or coauthored more than 200 technical papers. His current research interests include biometrics, pattern recognition, and computer vision. He is a fellow of the IAPR and serves as the Chair for IAPR
Technical Committee on Biometrics. He is an Associate Editor of the IEEE TRANSACTIONS ON BIOMETRICS, BEHAVIOR, AND IDENTITY SCIENCE.
\end{IEEEbiography}

\begin{IEEEbiography}[{\includegraphics[width=1in,height=1.25in,clip,keepaspectratio]{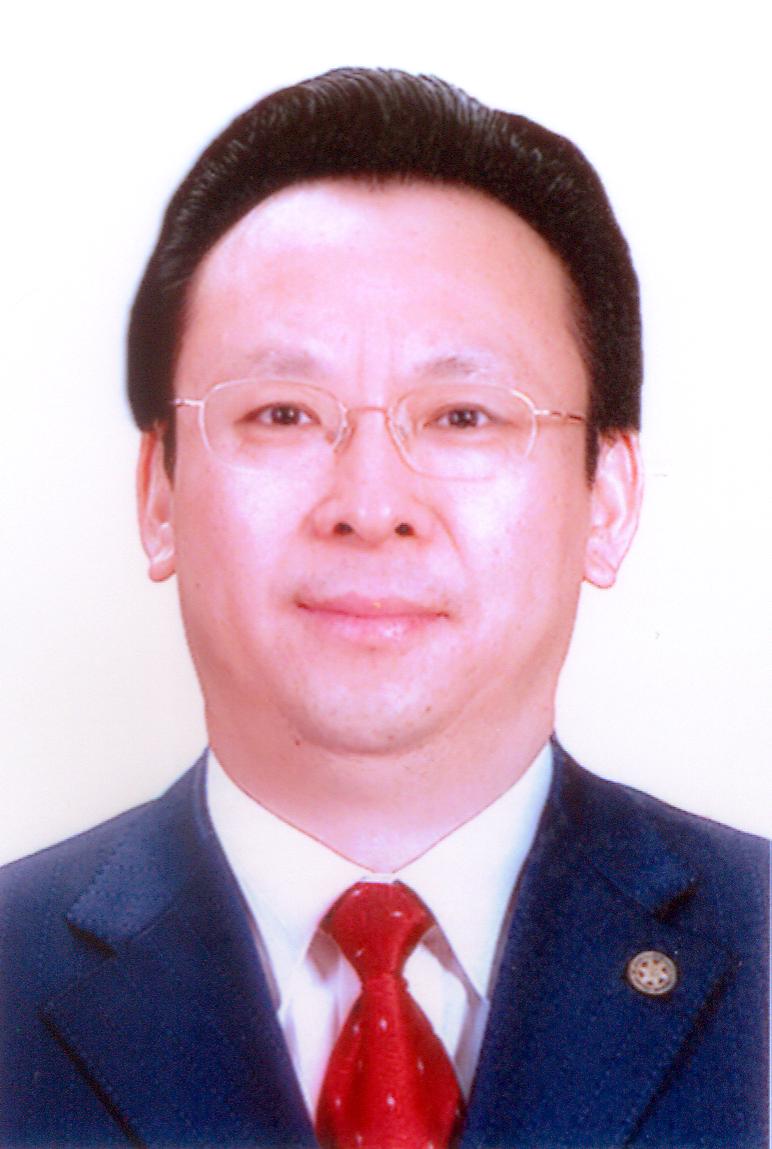}}]{Tieniu Tan}
received the BSc degree in electronic engineering from Xi’an Jiaotong University, China, in 1984, and the MSc and PhD degrees
in electronic engineering from Imperial College London, United Kingdom, in 1986 and 1989, respectively. He is currently a professor with the Center for Research on Intelligent Perception and Computing, NLPR, CASIA, China. He has published more than 450 research papers in refereed international journals and conferences in the areas of image processing, computer vision and pattern recognition, and has authored or edited 11 books. His research interests include biometrics, image and video understanding, information hiding, and information forensics. He is a fellow of the CAS, the TWAS, the BAS, the IEEE, the IAPR, the UK Royal Academy of Engineering, and the Past President of IEEE Biometrics Council. He is a fellow of the IEEE.
\end{IEEEbiography}

\vfill





\end{document}